\documentclass[11pt]{article}

\PassOptionsToPackage{dvipsnames}{xcolor}

\usepackage[]{acl}

\usepackage{times}
\usepackage{latexsym}
\usepackage[T1]{fontenc}
\usepackage[utf8]{inputenc}
\usepackage{microtype}
\usepackage{inconsolata}
\usepackage{graphicx}

\usepackage{amsmath}
\usepackage{amsfonts}
\usepackage{nicefrac}
\usepackage{booktabs}
\usepackage{multirow}
\usepackage{float}
\usepackage{wrapfig}
\usepackage{adjustbox}
\usepackage{longtable}
\usepackage{enumitem}
\usepackage{xurl}

\setlist[itemize]{noitemsep, nolistsep}
\makeatletter
\g@addto@macro\normalsize{%
  \setlength\abovedisplayskip{6pt plus 2pt minus 2pt}%
  \setlength\belowdisplayskip{6pt plus 2pt minus 2pt}%
  \setlength\abovedisplayshortskip{2pt plus 2pt}%
  \setlength\belowdisplayshortskip{4pt plus 2pt minus 2pt}}
\makeatother

\newcommand{\kvc}{\mathrm{KV}_c}
\newcommand{\kvp}{\mathrm{KV}_{c,\mathrm{evicted}}}
\newcommand{\thesistablestyle}{%
\scriptsize
\setlength{\tabcolsep}{3pt}%
\renewcommand{\arraystretch}{1.10}%
}
\newcommand{\selectionhighlight}[1]{%
\begingroup
\setlength{\fboxsep}{1pt}%
\colorbox{Goldenrod!75}{#1}%
\endgroup
}

\title{KV$^2$: A Self-Refining KV Cache}

\author{Johannes Wesch \phantom{\and} Danni Liu \phantom{\and} Jan Niehues \\
        Karlsruhe Institute of Technology, Germany \\
        \texttt{johannes.wesch@student.kit.edu}, \texttt{danni.liu@kit.edu}
        }

\hypersetup{
pdftitle={KV2: A Self-Refining KV Cache},
pdfsubject={Long-context LLM inference},
pdfkeywords={KV cache, cache compression, long-context inference, large language models},
hypertexnames=false,
}

\begin{document}
\maketitle

\begin{abstract}
    The memory footprint of the key-value (KV) cache constrains the practical use of long-context models, and it dominates cost when one prefilled context must later serve many different queries. In this reusable setting, query-agnostic compression trades cost against quality: lightweight estimators are cheap but less accurate, whereas full-context reconstruction scoring is more accurate yet reprocesses the entire prompt. We propose KV², a query-agnostic KV-cache compression method based on selective reconstruction. KV² first uses a lightweight proxy scorer to identify informative in-context tokens, then reprocesses only this subset to compute final eviction scores. On RULER, Needle-in-a-Haystack, and LongBench, KV²'s margin over baselines widens as the budget tightens: on RULER 16K at a 2\% KV-cache budget it improves the average score of Qwen3-8B over the next-best baseline by more than 40 percentage points, and on LongBench it attains the highest average at 2\%-5\% budgets by a wide margin (and remains ahead at 10\%), at lower compression-stage runtime and peak memory than full-context reconstruction. Reusable KV-cache compression thus does not require reprocessing the full context. Our code is available at \url{https://github.com/JohannesWesch/KVsquared}.
\end{abstract}

\begin{figure*}[!ht]
\centering
\includegraphics[width=0.65\linewidth]{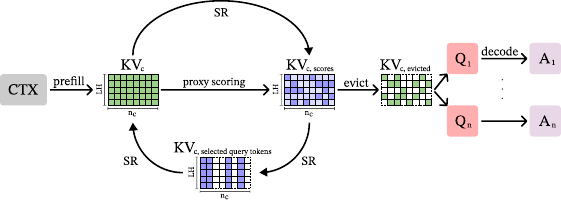}
\caption[KV² pipeline overview]{Overview of KV². After prefilling the context cache $\kvc$, the method computes proxy scores as a lightweight estimate of which in-context tokens are most informative. KV² then selects the query tokens $\mathrm{KV}_{c,\mathrm{selected\ query\ tokens}}$, runs \emph{Selective Reconstruction} (SR) on these selected queries to obtain cache scores, and evicts (in our implementation, masks) low-scoring KV entries before decoding reuses the compressed cache.}
%\vspace{-10pt} % replaced by global float spacing
\label{fig:introduction:overview}
\end{figure*}

\section{Introduction}
\label{sec:introduction}

Long-context large language models have considerably broadened the range of LLM applications, including analyzing long documents \cite{Bai2023,Zhang2024}, reasoning over larger code bases \cite{Bai2023,Zhang2024}, and maintaining extended conversational histories \cite{Zhang2024}. However, their practical use remains constrained by inference cost. A central reason is the key-value (KV) cache: once a context has been processed, the model must keep the resulting cached state available so that later tokens can attend to it without reprocessing. This cache grows linearly with context length, and increasingly dominates memory use during deployment.

Autoregressive inference separates into \emph{prefill} and \emph{decode} \cite{Kim2025,Feng2024} stages. 
During \emph{prefill}, the model processes the full prompt and computes keys and values for every layer, head, and token. 
For an $L$-layer Transformer decoder model with $H$ key-value heads and a context of $n_c$ tokens, this results in a cache $\kvc$, consisting of $L \cdot H \cdot n_c$ KV pairs. 
During \emph{decode}, new tokens are generated one at a time, 
and each query attends to $\kvc$ instead of reprocessing all previous tokens. 
Because $\kvc$ stores the exact keys and values produced by prefill, 
decoding from it yields the same output distribution as re-running prefill on the full context at every step \cite{Kim2025}.
%and each new query attends back to $\kvc$ instead of recomputing the prompt. %from scratch. 
%The cache must be a sufficient statistic for the prompt: conditional generation on $\kvc$ yields the same predicted distribution as conditioning on the original context itself \cite{Kim2025}.
% Let $\mathcal{V}$ be the vocabulary and $\ct$ be the space of token sequences. 
% Following the notations from Kim et al. \cite{Kim2025}, an autoregressive Transformer-based decoder model with $L$ layers, each with $H$ key-value heads, maps a context $c \in \ct$ to a distribution $\lm(\cdot \mid c)$ over $\mathcal{V}$. 
% Prefilling a context of length $n_c$ yields a key-value cache $\kvc$ comprising $L \cdot H \cdot n_c$ KV pairs, which serves as a sufficient statistic for $c$: by construction, $\lm(\cdot \mid \kvc) = \lm(\cdot \mid c)$.
%KV-cache compression then seeks a smaller cache $\kvc'$ that approximates this equivalence.
%let $\lm:\ct\rightarrow\ct$ be an autoregressive Transformer-based LLM with $L$ layers and $H$ KV heads. 
%Given a context $c \in \ct$ tokenized into $n_c$ tokens, prefill generates a cache $\kvc$ containing $L \times H \times n_c$ KV pairs, and conditioned generation is written as $\lm(\cdot \mid \kvc)$ \cite{Kim2025}. 

% This equivalence comes at a cost that scales linearly in $n_c$,
% where each additional token contributes $L \cdot H$ new KV pairs to the resident cache.
% At long contexts, this becomes a bottleneck:
Maintaining this equivalence is what makes the cache expensive. 
Every context token contributes $L\cdot H$ new KV pairs to $\kvc$, 
and the resident footprint grows accordingly.
% This notation already makes the dominant scaling explicit: every additional context token contributes $L \times H$ new KV pairs, so the resident cache grows linearly with $n_c$ until the request finishes or compression is applied \cite{Kim2025,Chari2025}. In practice, this quickly becomes a systems bottleneck: 
A 100K-token prompt on Qwen-2.5-32B already requires about 26\,GB of KV memory \cite{Chari2025}, and a one-million-token cache for a 70B-scale model would require roughly 320\,GB of GPU memory \cite{Devoto2025a}. %Prefill is compute-bound on the full prompt,
%expensive because the model must process the whole prompt, while 
Decoding is typically memory-bandwidth-bound, 
since each generation step rereads the cache from memory \cite{Chari2025,Shazeer2019,Yang2024}. Architectural choices such as Multi-Query Attention (MQA) \cite{Shazeer2019}, Grouped Query Attention (GQA) \cite{Ainslie2023}, and Multi-Head Latent Attention (MLA) \cite{deepseekv2,transMLA} reduce the baseline cache, but they do not remove the underlying KV-cache bottleneck. %\cite{Shazeer2019,}.
KV-cache compression is therefore essential for long-context deployment, 
where the goal is to replace $\kvc$ with a smaller cache $\kvp$ that approximately preserves the equivalence above.

Beyond these architectural reductions, prior work 
falls into two broad categories. 
The first improves attention kernels and serving systems, e.g., FlashAttention or PagedAttention \cite{Dao2023,Kwon2023},  reducing realized memory and compute without altering what the cache stores. 
%either improves attention kernels and serving systems, for example through FlashAttention or PagedAttention \cite{Dao2023,Kwon2023}, 
The second compresses the cache itself by evicting low-importance tokens \cite{Xiao2023,Zhang2023}, 
%or compresses the cache itself by evicting low-importance tokens \cite{Xiao2023,Zhang2023}, 
retaining evicted states in low precision \cite{Yang2024}, or pruning cache channels \cite{Xu2024}. Many recent eviction methods are \emph{query-aware}: they keep entries based on a known downstream question or on a \textit{proxy} for the next generation step \cite{Li2024,Wang2025,Liu2025}. This is effective when the next use of the cache is known, but less so when one compressed cache must remain useful across many later queries, e.g., for long-context question-answering or persistent agent contexts.
Another line of work targets \emph{query-agnostic} compression \cite{Devoto2025a, Kim2025, Chari2025, Kim2026}, which scores entries without reference to any specific downstream query, and is the focus of this work. In our experiments, the question is appended only after compression and is never part of the compressed context. The context may, however, include the task instruction and few-shot demonstrations, so our setting is more precisely \emph{task-aware, query-agnostic} (\S\ref{sec:experiments:setup}).

%We focus instead on the \emph{reusable} cache-compression setting emphasized by recent query-agnostic work \cite{Kim2025,Chari2025}, where one prefilled context must remain useful across multiple later queries, as in repeated document question answering or multi-turn assistants. Formally, if $\kvp \subseteq \kvc$ denotes a compressed version of the prefilled cache, the reusable-cache objective is to preserve the model's behavior across future queries rather than only one anticipated continuation:
%\begin{equation}
%\lm(q \mid \kvp) \approx \lm(q \mid \kvc),
%\qquad \forall q \in \ct.
%\label{eq:introduction:reusable-objective}
%\end{equation}

We introduce KV\textsuperscript{2}, a query-agnostic KV-cache compression method aimed at the reusable setting described above\footnote{The superscript in KV\textsuperscript{2} reflects the method's two consecutive KV-based stages: proxy-based token selection followed by reconstruction-based final scoring.},
in which one prefilled cache must serve many later queries while sharing device memory with other requests.
This setting motivates retention ratios well below 10\% of the original cache, lower than those typically studied in prior work \cite{Gao2024,Srivatsa2024,Abhyankar2024,kvpress2025}.
Throughout, budgets denote nominal retention ratios, i.e., the fraction of KV entries that remain visible to attention. Our implementation realizes them by masking, so converting them into resident-memory savings requires physical cache compaction (see Limitations).
KV\textsuperscript{2} is motivated by a simple observation: 
%reconstruction-based scoring is typically more accurate than lightweight proxies, but computing it across the full context wastes compute at the tight budgets where compression matters most. %stronger context-centered scoring can be more robust than lightweight proxy methods alone, but applying that signal uniformly across the full context is both expensive and not always the best use of compression-time compute. 
long-context models do not use all positions equally, and 
adding context can dilute the model's access to the tokens that genuinely impact downstream behavior
%more context can increase the amount of content the model must reason over rather than improving access to the truly relevant evidence 
\cite{Liu2023a}. 
At such tight budgets, 
most positions are either trivial to retain (e.g., attention sinks) or trivial to evict (e.g., filler words and punctuation).
%a broad whole-context pass can therefore still spend capacity on filler words, punctuation, or semantically unimportant sink tokens,
The eviction decision therefore depends on a sparse set of informative anchors
%rather than on the sparse anchors that determine which cache entries matter most downstream 
\cite{Ainslie2023b,Xiao2023,Zhang2023}.
Yet, existing reconstruction-based methods score every cached entry uniformly,
by recomputing attention output over the full context and measuring each position's contribution, which in turn pays the same cost for \textit{both} trivial and informative tokens. 
%Even though only a small fraction of tokens carry this essential signal, 
%a full-context reconstruction pass scores every position uniformly.
Instead, we show that scoring a small, well-chosen set of in-context tokens often suffices to identify the cached entries that preserve the model's behavior on the original context.

KV\textsuperscript{2} turns this intuition into a two-stage selective-reconstruction pipeline (Figure~\ref{fig:introduction:overview}). %shows, the method first uses 
A lightweight proxy scorer first identifies a small set of informative context tokens.
These tokens are then used as reconstruction queries against the full cache to compute the final eviction scores. This concentrates the expensive scoring pass on the dependencies most likely to matter under severe compression, which both stabilizes the ranking and lowers compression-time cost. This two-stage design can swap the proxy scorer, although proxy quality determines how much of the gain is realized and KeyDiff is our recommended default (\S\ref{sec:analysis:proxy}). It also supports iterative self-refinement when additional compression-time compute is available (\S\ref{sec:analysis:iterative-refinement}).

% We evaluate KV\textsuperscript{2} on \texttt{Llama-3.1-8B-Instruct} and \texttt{Qwen3-8B} across controlled retrieval diagnostics and broader long-context tasks, including RULER 4K/16K, Needle-in-a-Haystack, and LongBench. %Our main findings are threefold. First, KV\textsuperscript{2} is strongest precisely when reusable compression becomes genuinely selective rather than merely approximate, and we make this severe regime explicit by evaluating down to 2\% and even 1\% KV size, which is rarely reported in prior work: on RULER, it remains competitive through 90\% eviction, and at more aggressive budgets the gap widens quickly; at a 95\% eviction ratio on standalone Needle-in-a-Haystack, it is likewise the most reliable method for recovering the planted fact across deep insertion depths. Second, these gains carry over to broader long-context workloads: on LongBench, KV\textsuperscript{2} achieves the best average results across tight 2\%--10\% budgets on both models. Third, compression-stage runtime and peak-memory measurements show that single-pass KV\textsuperscript{2} offers a stronger quality-efficiency operating point than applying the same second-stage scoring pass to every token.
Our main contributions are threefold:
\begin{itemize}[nolistsep,leftmargin=*]
    \item \textbf{Selective reconstruction for reusable KV-cache compression.} We introduce KV\textsuperscript{2}, a task-aware, query-agnostic two-stage method that uses a lightweight proxy to select a small set of in-context tokens and reprocesses only this subset to compute final eviction scores (\S\ref{sec:method}).
    \item \textbf{Strong performance under extreme compression.} We evaluate nominal KV-cache budgets down to 1-2\%, a regime motivated by the reusable setting and rarely studied in prior work, and find increasing gains of KV\textsuperscript{2} over baselines as the budget tightens (\S\ref{sec:experiments}).%sec:experiments:ruler16k}\ref{sec:experiments:needle},\ref{sec:experiments:longbench})
    \item \textbf{Favorable efficiency profile.} Compression-stage runtime and peak-memory measurements show that, at the compression stage, KV\textsuperscript{2} offers a better quality–efficiency trade-off than applying full-context reconstruction scoring over every token (\S\ref{sec:analysis:single-pass}). Decode-time savings require physical compaction and are not measured here.
\end{itemize}

\vspace{-3pt}
\section{Related work}
\label{sec:related-work}

%We review prior KV-cache compression methods through the lens most relevant to KV\textsuperscript{2}: how methods decide which cached states to retain and how they define token importance.
\noindent
\textbf{Early training-free eviction.} %Early KV-cache compression methods showed that attention mass is highly uneven, so retaining every cached token is often unnecessary.
StreamingLLM keeps recent tokens together with a small set of initial ``attention sink'' tokens \cite{Xiao2023}, and H$_2$O emphasizes heavy hitters that repeatedly attract useful attention \cite{Zhang2023}. These early works established several themes that reappear throughout the literature. First, token utility is highly unequal. Second, recency is important but insufficient on its own. Third, substantial compression can often be achieved without retraining by exploiting signals already present in the inference process, especially attention patterns.

\noindent
\textbf{Anticipating future decoding.} Later methods increasingly score tokens by how likely they are to matter for future decoding. SnapKV uses an observation window near the end of the prompt \cite{Li2024}, LAQ generates pseudo queries \cite{Wang2025}, and Judge Q learns surrogate query tokens \cite{Liu2025}. These methods can work well when compression is guided by the continuation the model is about to decode, but they are less suitable for reusable caches that must support multiple later questions. For that setting, Expected Attention estimates token importance under a distribution of possible future queries \cite{Devoto2025a}, while KeyDiff uses key geometry as a query-agnostic proxy \cite{Park2025}. Compactor ranks tokens by approximate leverage scores and calibrates the compression ratio per context \cite{Chari2025}. All three are closer to our reusable setting because they avoid specializing the cache to one anticipated continuation.

\noindent
\textbf{Context reconstruction.}
%The closest prior line of work to ours is context reconstruction.
Context reconstruction changes the objective altogether. Instead of asking which cached tokens one future continuation is likely to use, it asks whether the model needs them to reproduce the prompt itself. KVzip introduced this viewpoint \cite{Kim2025}, aligning the scoring objective closely with reusable caches but incurring the cost of broad reconstruction. Subsequent work keeps this reconstruction-based perspective while trying to reduce memory and compute. Fast KVzip and KVzap distill reconstruction-derived importance into lightweight trained predictors \cite{Kim2026,Jegou2026}, which avoids a reconstruction pass at compression time but requires additional training. These methods are the most direct precursors to KV\textsuperscript{2} and share its reusable, query-agnostic objective. KV\textsuperscript{2} takes the complementary, training-free route: it retains an actual reconstruction pass, but makes that pass selective by shrinking the reconstruction query set itself. The two directions compose naturally, since a distilled gate could serve as the Stage~1 proxy of KV\textsuperscript{2} (\S\ref{sec:method}).\looseness=-1

\vspace{-3pt}
\section{Method}
\label{sec:method}

We now describe the overview from Figure~\ref{fig:introduction:overview} in detail. KV\textsuperscript{2} is a two-stage compression procedure. Starting from a prefilled cache $\kvc$, it
processes the prompt chunk by chunk. 
In \emph{Stage 1 (query selection)}, the goal is to identify a small set of in-context tokens that, when used as queries, will most informatively rank the cached entries.
KV\textsuperscript{2} produces this selection with a lightweight proxy scorer.
In \emph{Stage 2 (cache scoring)}, the goal is to assign every cached entry a final eviction score. KV\textsuperscript{2} reprocesses only the queries selected in Stage 1 against the full cache to produce the scores.
%Within each chunk, a proxy scorer selects a small set of reconstruction queries (Stage 1), 
%which are then reprocessed against the full cache to produce final eviction scores (Stage 2).
%it first computes lightweight proxy scores over the context and uses them to select a small set of informative in-context tokens. These selected tokens are then replayed against the full cache in a second-stage \emph{Selective Reconstruction} (SR) pass to produce the final cache scores used for eviction. 
The core design choice is therefore not \textit{whether} to reconstruct, but \textit{which} queries to use for reconstruction.
After outlining the main method, we describe an optional iterative refinement and analyze complexity.

\begin{figure}[t]
\centering
\includegraphics[width=0.58\linewidth]{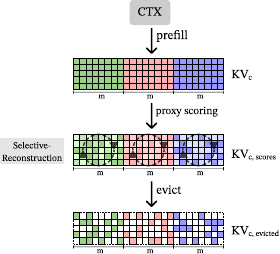}
\caption[Chunked selective reconstruction]{Chunked processing in KV$^2$. The context is split into chunks. Each chunk proposes its own reconstruction queries, and the resulting local scores are merged into one global eviction decision.}
\label{fig:method:chunking}
\end{figure}

\paragraph{Chunking.}
Following KVzip \cite{Kim2025}, KV\textsuperscript{2} does not score the prompt with a single global reconstruction set.
%construct one global replay set over the entire prompt. 
Instead, it retains a small set of \textit{sink} tokens at the start of the prefix verbatim and partitions the rest of the cache, i.e., the \textit{compressible portion} into chunks:
\begin{equation}
C_1,\dots,C_M,
\qquad
|C_r| \le m,
\label{eq:method:chunks}
\end{equation}
where $m$ is the chunk size.
%is actually eligible for eviction; a small protected prefix of sink tokens is excluded from chunking and retained throughout. 
As illustrated in Figure~\ref{fig:method:chunking}, this structure keeps reconstruction-based scoring tractable on long prompts while allowing different context regions to propose different reconstruction 
queries before their local scores are merged into one global eviction decision.

\paragraph{Proxy scoring and query selection.} 
Stage~1 needs a per-token signal that is much cheaper than full reconstruction yet informative enough to surface promising reconstruction queries.
% The first stage of KV\textsuperscript{2} needs a proxy signal that is much cheaper than full reconstruction, yet informative enough to identify promising reconstruction queries. 
By default, we use the KeyDiff score \cite{Park2025}. Alternative proxy scorers are compared in Section~\ref{sec:analysis:proxy}. The proxy only proposes reconstruction queries and never determines eviction directly, which is always decided by the Stage~2 scores (Eq.~\ref{eq:method:kvsquared}).
%In the default configuration used here, that signal is the efficient KeyDiff score \cite{Park2025}, although we compare multiple alternative proxy scorers later in Section~\ref{sec:analysis:proxy}. 
For each token $i \in C_r$, layer $\ell$, and KV head $h$, let $d_{i,\ell,h}$ denote the resulting layer- and head-specific KeyDiff score. KV\textsuperscript{2} aggregates the scores to the token level by averaging over layers and KV heads:
\begin{equation}
p_i
=
\frac{1}{L H}
\sum_{\ell=1}^{L}\sum_{h=1}^{H} d_{i,\ell,h},
\qquad
i \in C_r.
\label{eq:method:proxy}
\end{equation}
Averaging proved stronger than a max-based aggregation in our initial experiments.
Intuitively, tokens that appear distinctive across many layers and heads accrue larger scores. 
%This averaging, which proved stronger than a max-based aggregation in our initial experiments,  produces a token-level proxy score rather than a headwise eviction decision. 
%Tokens that appear distinctive across many layers or KV heads receive larger proxy scores and are therefore more likely to be reconstructed. %Given these proxy scores, %KV\textsuperscript{2} selects the highest-scoring positions as an ordered replay set
KV\textsuperscript{2} then selects the $n_q$ highest-scoring positions in $C_r$ as an ordered reconstruction set:
\begin{equation}
\begin{gathered}
Q_r = (q_{r,1},\dots,q_{r,n_q}),
\quad
q_{r,1} < \cdots < q_{r,n_q}, \\
n_q = \max\!\left(\lfloor \rho |C_r| \rfloor, 1\right),
\end{gathered}
\label{eq:method:query-selection}
\end{equation}
where $\rho$ is the query-selection ratio. %The positions $q_{r,1},\dots,q_{r,n_q}$ are the $n_q$ highest-scoring tokens in $C_r$. 
The selected positions do not need to form a contiguous span in $C_r$. KV² treats them as a sparse set of probe queries: during reconstruction, they are appended after the prefilled cache $\kvc$, 
assigned position IDs immediately following the prefilled context, %receive fresh positional encodings, 
and query the existing cache under the causal mask rather than being reinserted at original locations.
We inherit this design from KVzip \cite{Kim2025}: reinserting a token at its original position would collapse its reconstruction query onto the query it already formed during prefill and remove the reconstruction signal.

\paragraph{Selective reconstruction.} For one chunk $C_r$, KV\textsuperscript{2} uses only the selected context tokens in $Q_r=(q_{r,1},\dots,q_{r,n_q})$ as reconstruction queries. These reconstruction tokens are appended after the prefilled cache $\kvc$ and processed in a new causal forward pass with access to the full cache, so their hidden states and query vectors have whole-context conditioning and can draw on evidence from anywhere in the prompt. 

\begin{figure}[!ht]
\centering
\includegraphics[width=1.05\linewidth]{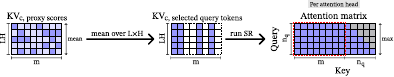}
\caption[KV² selective reconstruction]{Detailed view of KV² inside one chunk. Proxy scores first select the ordered reconstruction set $Q_r$. These reconstruction queries are then encoded against the shared cache, and the final score for a cached position is the maximum attention it receives during selective reconstruction.}
%\vspace{-10pt} % replaced by global float spacing
\label{fig:method:attention-matrix}
\end{figure}

Unlike the global conditioning for the reconstruction tokens, the final score extraction is restricted to the current chunk (Figure~\ref{fig:method:attention-matrix}). 
Each reconstruction query's attentions are only used to score positions in $C_r$.
Because attention is renormalized within each chunk during score extraction, the shared appended position IDs do not systematically shift the budget toward chunks closer to the end of the prompt.
This \textit{global-conditioning, local-scoring} split is the main design choice of KV\textsuperscript{2}. Concretely, for one layer $\ell$ and one KV head $h$, let $a_{ji}^{(\ell,h)}$ denote the attention paid by the $j$-th reconstruction query in $Q_r$ to cached position $i \in C_r$ (under grouped-query attention, the maximum over the query heads that share KV head $h$). KV\textsuperscript{2} then assigns the chunk-local score:
\begin{equation}
s_{i,\ell,h}
=
\max_{j=1,\dots,n_q} a_{ji}^{(\ell,h)},
\qquad
i \in C_r.
\label{eq:method:kvsquared}
\end{equation}
The max ensures a cached position is retained whenever \emph{any} reconstruction query depends on it strongly, rather than requiring agreement across all queries.\footnote{Each selected token can also attend to its own original cache entry, so Eq.~\ref{eq:method:kvsquared} could in principle mostly reward self-matches. Since $\rho$ equals the retained budget by default, KV\textsuperscript{2} would then essentially retain its proxy's top tokens and track KeyDiff at equal budget. It does not: on RULER 4K at a 2\% budget, KV\textsuperscript{2} reaches 53.36/49.05 (Llama/Qwen) versus 23.22/10.37 for KeyDiff (Appendix~\ref{sec:appendix:ruler4k}).}
%A cached position therefore receives a high score if at least one reconstructed token from the same chunk depends on it strongly during reconstruction. 
%The max ensures a cached position is retained whenever any reconstruction query depends on it, rather than all of them.
After all chunks have been processed, KV\textsuperscript{2} gathers the local scores into a global score tensor, pins the sink positions at the maximal score so they remain in the cache, and evicts the lowest-scoring entries under the target compression budget (realized as head-wise masks in our implementation, see \S\ref{sec:experiments:setup}).
%the local scores are written back into one global score tensor, 
%the protected sink positions are assigned maximal score for guaranteed retention,
%a small set of initial sink positions is protected by assigning them maximal score from the outset, 
%and the lowest-scoring entries are evicted under the target compression budget.

\paragraph{Iterative refinement.} The same two-stage construction also defines a self-refinement procedure. 
Since Stage~2 produces a per-token score for each position in $C_r$, at the same granularity as the Stage~1 proxy signal $p_i$, the output of one pass can directly replace the proxy in the next.
%Because selective reconstruction outputs chunk-local score vectors, 
%the scores from one pass can serve as the proxy signal for the next.
%the scores produced by one pass can become the proposal signal for the next pass within the same chunk. 
Single-pass KV\textsuperscript{2} therefore corresponds to the first refinement round, and we adopt it as the default because it offers the strongest quality-efficiency trade-off.
The multi-pass variant is studied in Section~\ref{sec:analysis:iterative-refinement} as an optional extension.\looseness=-1
%; later analysis studies the multi-pass variant as an optional extension.

\paragraph{Complexity analysis.} 
The proxy pass added before reconstruction is linear in the prompt length, so the reconstruction forward pass remains the dominant cost of KV\textsuperscript{2}. %KV\textsuperscript{2} adds only a proxy pass linear with the prompt length before reconstruction, so the dominant cost remains reconstructing selected queries against the shared cache. 
Relative to KVzip, 
which reconstructs over every token in a chunk, 
%which reconstructs the full chunk, 
KV\textsuperscript{2} reconstructs only a proxy-selected fraction $\rho$, so its reconstruction-stage runtime and peak activation footprint scale with $\rho$ rather than with full-chunk reconstruction.
%; runtime and peak activation footprint therefore scale with the replay ratio rather than with full-chunk replay. 
Score extraction is cheaper because it remains chunk-local: after the reconstruction forward pass, KV\textsuperscript{2} recomputes attention only on the reduced $n_q \times (m + n_q)$ matrix shown in Figure~\ref{fig:method:attention-matrix}. The reconstruction forward can still use standard optimized causal-attention kernels, while the final score extraction is a smaller chunk-local postprocessing step. Appendix~\ref{sec:appendix:complexity} gives the fuller asymptotic derivation.\looseness=-1

\vspace{-3pt}
\section{Experiments}
\label{sec:experiments}
After introducing the experimental setup (\S\ref{sec:experiments:setup}), we present experiments on RULER (\S\ref{sec:experiments:ruler16k}), Needle-in-a-Haystack (\S\ref{sec:experiments:needle}), and LongBench (\S\ref{sec:experiments:longbench}).
\vspace{-3pt}
\subsection{Experimental setup}
\label{sec:experiments:setup}
\noindent
\textbf{Data and models.} We use three datasets from controlled retrieval stress tests and realistic long-context tasks: RULER 4K/16K \cite{Hsieh2024}, LongBench \cite{Bai2023}, and a Needle-in-a-Haystack diagnostic built from Paul Graham essays \cite{Hsieh2024,DevotoPaulGrahamEssays}. We mainly use two backbones, \texttt{Llama-3.1-8B-Instruct} \cite{Dubey2024} and \texttt{Qwen3-8B} \cite{Yang2025b,Qwen3ModelCard2025}.
Results from \texttt{Qwen2.5-7B-Instruct-1M} \cite{Yang2025} and InfiniteBench \cite{zhang-etal-2024-bench} are in the appendices. InfiniteBench is evaluated on a single 10\% subsample and excludes \texttt{math\_find}, whose prompt template embeds the question. Inputs longer than a model's context window are truncated to fit it (128K tokens for \texttt{Llama-3.1-8B-Instruct}).\looseness=-1

\noindent
\textbf{Implementation details.} Unless stated otherwise, KV$^2$ uses a chunk size of 4K tokens, KeyDiff as the first-stage proxy scorer, and by default sets the query-selection ratio $\rho$ to the final retained KV-cache ratio. Tighter compression budgets therefore use smaller reconstruction sets.
For every method, compression is applied only to the forward pass over the context. The question is appended after compression, so all caches are query-hidden. The compressed context does retain the task instruction and, for RULER \texttt{vt}/\texttt{cwe}/\texttt{fwe} and LongBench TREC/TriviaQA/SAMSum, the few-shot demonstrations.
% TODO (paJ1 C1): add an appendix with the exact prompt split per benchmark and reference it here.
%More aggressive compression therefore uses smaller replay sets under this default.
All runs use PyTorch and Hugging Face Transformers with FlashAttention-2 \cite{Dao2023}, are built on NVIDIA's \texttt{kvpress} framework\footnote{\texttt{kvpress} realizes head-wise methods (KVzip, Expected Attention with Ada-KV budget allocation \cite{Feng2024}, and KV$^2$) through masking, whereas KeyDiff retains the same number of entries in every head.} \cite{kvpress2025}, and run on
%. The runs were carried out 
a single NVIDIA A100 80\,GB or H100 94\,GB GPU.
All methods are compared at matched nominal budgets.

\noindent
\textbf{Baselines.}
In the main comparisons, we compare to three state-of-the-art query-agnostic methods that cover the main families of approaches: lightweight proxy scoring (KeyDiff \cite{Park2025}), expected future-query scoring (Expected Attention \cite{Devoto2025a}), and reconstruction-based compression (KVzip \cite{Kim2025}).
We do not compare against Fast KVzip and KVzap, which require training and are complementary to KV$^2$ (a distilled gate could serve as its Stage~1 proxy), so KVzip serves as the reference for reconstruction-based scoring. Compactor calibrates a separate compression ratio for each context, which does not fit our fixed-budget sweep.
For the choice of first-stage proxy scorer inside KV$^2$, we compare KeyDiff, Knorm \cite{Devoto2024}, CUR \cite{Sengupta2025}, 
and StreamingLLM \cite{Xiao2023}, while keeping the same second-stage selective reconstruction fixed.\looseness=-1

\begin{table*}[!ht]
\centering
\thesistablestyle
\caption[LongBench results across KV retention ratios]{LongBench results across KV-cache budgets for \texttt{Llama-3.1-8B-Instruct} and \texttt{Qwen3-8B}. For both, KV² gives the strongest average performance at tighter budgets, with KVzip becoming closer at 10\% retention.}
\label{tab:evaluation:longbench}

\resizebox{\textwidth}{!}{%
\begin{tabular}{@{}lccccccccccccccccc@{}}
\toprule
\multirow{2}{*}{\textbf{Method}} 
    & \multicolumn{3}{c}{\textbf{Single-doc QA}}
    & \multicolumn{3}{c}{\textbf{Multi-doc QA}} 
    & \multicolumn{3}{c}{\textbf{Summ.}}
    & \multicolumn{3}{c}{\textbf{Few-shot / Cls.}}
    & \multicolumn{2}{c}{\textbf{Synth. Count}}
    & \multicolumn{2}{c}{\textbf{Code}}
    & \multirow{2}{*}{\textbf{Avg.}} \\
\cmidrule(lr){2-4} \cmidrule(lr){5-7} \cmidrule(lr){8-10} \cmidrule(lr){11-13} \cmidrule(lr){14-15} \cmidrule(lr){16-17}
    & NrtvQA & Qasper & MFQA-en
    & Hotpot & 2Wiki & MuSiQ
    & GovRep & QMSum & MultiN
    & TREC & TrivQA & SAMSum
    & PCount & PR-en
    & LCC & RB-P
    & \\
\midrule

%% ============================================================
%% Llama-3-8B-Instruct
%% ============================================================
\multicolumn{18}{c}{\textbf{Llama-3.1-8B-Instruct}} \\
\midrule

Full 
    & 30.21 & 45.53 & 55.01
    & 56.01 & 46.65 & 31.28
    & 35.13 & 25.28 & 27.25
    & 73.00 & 91.64 & 43.80
    & 8.91 & 99.50
    & 63.38 & 56.64
    & 49.33 \\

\midrule
\multicolumn{18}{c}{\textit{KV Size = 2\%}} \\
\midrule

KeyDiff 
    & 17.24 & 12.11 & 16.71
    & 26.07 & 20.44 & 5.06
    & 21.65 & 18.36 & 10.99
    & 17.00 & \textbf{91.34} & 36.60
    & 6.00 & 25.50
    & 12.66 & 48.75
    & 24.16 \\

KVzip 
    & 12.54 & 11.10 & 18.98
    & 24.80 & 21.78 & 8.69
    & 11.39 & 17.40 & 7.36
    & 5.00 & 87.47 & 32.31
    & 2.50 & 4.00
    & 18.20 & \textbf{51.75}
    & 20.95 \\

KV²
    & \textbf{23.11} & \textbf{14.43} & \textbf{24.23}
    & \textbf{43.12} & \textbf{25.42} & \textbf{14.91}
    & \textbf{25.05} & \textbf{20.99} & \textbf{20.07}
    & \textbf{31.50} & 86.89 & \textbf{38.07}
    & \textbf{10.87} & \textbf{76.50}
    & \textbf{33.85} & 48.41
    & \textbf{33.59} \\

\midrule
\multicolumn{18}{c}{\textit{KV Size = 5\%}} \\
\midrule

KeyDiff 
    & 22.55 & 13.26 & 23.73
    & 31.26 & 20.76 & 9.86
    & 24.29 & 20.02 & 15.63
    & 28.00 & \textbf{91.99} & \textbf{38.32}
    & 10.00 & 54.00
    & 16.25 & 46.00
    & 29.12 \\

KVzip 
    & 22.47 & 16.45 & 36.44
    & 39.06 & 20.62 & 10.57
    & 22.04 & 21.31 & 17.26
    & 10.50 & 90.96 & 35.24
    & 4.50 & 23.00
    & 28.57 & \textbf{52.35}
    & 28.21 \\

KV²
    & \textbf{28.20} & \textbf{28.07} & \textbf{37.29}
    & \textbf{47.66} & \textbf{34.90} & \textbf{21.58}
    & \textbf{27.51} & \textbf{22.99} & \textbf{23.13}
    & \textbf{44.00} & 87.06 & 37.86
    & \textbf{10.75} & \textbf{98.00}
    & \textbf{41.31} & 48.20
    & \textbf{39.91} \\

\midrule
\multicolumn{18}{c}{\textit{KV Size = 10\%}} \\
\midrule

KeyDiff 
    & 27.23 & 16.57 & 35.79
    & 40.89 & 23.72 & 16.86
    & 26.21 & 21.20 & 18.98
    & 40.50 & \textbf{91.74} & \textbf{39.47}
    & 9.54 & 89.50
    & 21.29 & 45.12
    & 35.29 \\

KVzip 
    & \textbf{29.92} & \textbf{40.92} & \textbf{54.33}
    & 51.80 & 37.65 & \textbf{27.60}
    & \textbf{32.14} & 23.64 & \textbf{25.45}
    & 44.50 & 90.12 & 37.93
    & 10.50 & 61.00
    & 42.08 & \textbf{51.21}
    & 41.30 \\

KV² 
    & 29.40 & 36.50 & 48.25
    & \textbf{54.91} & \textbf{44.79} & 26.35
    & 30.45 & \textbf{24.10} & 24.85
    & \textbf{51.50} & 85.06 & 38.70
    & \textbf{11.17} & \textbf{99.00}
    & \textbf{44.50} & 47.98
    & \textbf{43.59} \\

\midrule
%% ============================================================
%% Qwen3-8B
%% ============================================================
\multicolumn{18}{c}{\textbf{Qwen3-8B}} \\
\midrule

Full
    & 29.76 & 43.66 & 55.70
    & 62.87 & 47.53 & 35.31
    & 33.33 & 24.56 & 24.79
    & 40.50 & 90.56 & 40.03
    & 10.00 & 91.18
    & 64.97 & 59.96
    & 47.17 \\

\midrule
\multicolumn{18}{c}{\textit{KV Size = 2\%}} \\
\midrule

KeyDiff
    & 5.86 & 3.60 & 18.91
    & 9.23 & 21.16 & 3.75
    & 8.88 & 17.56 & 4.65
    & 2.50 & 72.21 & 27.64
    & 3.75 & 4.38
    & 8.05 & 53.29
    & 16.59 \\

KVzip
    & 6.60 & 5.76 & 16.04
    & 6.15 & 8.16 & 4.35
    & 11.89 & 16.38 & 8.44
    & 1.50 & 58.72 & 31.82
    & 0.50 & 4.50
    & 15.94 & \textbf{60.87}
    & 16.10 \\

KV²
    & \textbf{24.79} & \textbf{14.65} & \textbf{27.06}
    & \textbf{37.90} & \textbf{28.93} & \textbf{17.79}
    & \textbf{24.05} & \textbf{19.94} & \textbf{13.16}
    & \textbf{14.75} & \textbf{89.67} & \textbf{36.35}
    & \textbf{5.50} & \textbf{67.42}
    & \textbf{44.69} & 60.80
    & \textbf{32.97} \\

\midrule
\multicolumn{18}{c}{\textit{KV Size = 5\%}} \\
\midrule

KeyDiff
    & 8.01 & 2.17 & 19.77
    & 12.47 & 19.81 & 5.56
    & 12.17 & 18.67 & 6.03
    & 4.50 & 75.99 & 29.06
    & 4.05 & 5.00
    & 7.73 & 51.44
    & 17.65 \\

KVzip
    & 12.89 & 9.65 & 26.65
    & 12.70 & 10.16 & 7.77
    & 20.76 & 20.89 & 10.77
    & 14.50 & 77.67 & 32.95
    & 3.50 & 13.50
    & 25.22 & \textbf{62.14}
    & 22.61 \\

KV²
    & \textbf{25.73} & \textbf{22.86} & \textbf{32.79}
    & \textbf{44.42} & \textbf{34.79} & \textbf{20.22}
    & \textbf{27.55} & \textbf{21.93} & \textbf{17.03}
    & \textbf{35.00} & \textbf{88.94} & \textbf{37.28}
    & \textbf{8.68} & \textbf{85.20}
    & \textbf{49.95} & 60.15
    & \textbf{38.28} \\

\midrule
\multicolumn{18}{c}{\textit{KV Size = 10\%}} \\
\midrule

KeyDiff
    & 9.75 & 8.53 & 22.97
    & 14.43 & 23.54 & 6.95
    & 16.15 & 19.70 & 8.19
    & 7.00 & 78.30 & 31.79
    & 7.61 & 15.17
    & 12.67 & 49.74
    & 20.78 \\

KVzip
    & 25.58 & \textbf{36.61}
    & \textbf{49.80}
    & 53.09 & 39.79 & 23.83
    & \textbf{30.98} & 22.80 & \textbf{23.01}
    & 56.00 & 88.74 & 36.45
    & \textbf{11.00} & 58.42
    & 47.23 & \textbf{62.57}
    & 41.62 \\

KV²
    & \textbf{25.94} & \textbf{33.60} & 41.89
    & \textbf{54.18} & \textbf{40.79} & \textbf{27.43}
    & 30.37 & \textbf{23.42} & 20.71
    & \textbf{59.25} & \textbf{90.09} & \textbf{38.61}
    & 9.60 & \textbf{90.38}
    & \textbf{56.44} & 59.58
    & \textbf{43.89} \\

\bottomrule
\end{tabular}
}

\end{table*}

%% ============================================================
%% Column to data_dir mapping for evaluation:
%% ============================================================
%% NrtvQA   -> narrativeqa
%% Qasper   -> qasper
%% MF-en    -> multifieldqa_en
%% HotpotQA -> hotpotqa
%% 2WikiMQA -> 2wikimqa
%% Musique  -> musique
%% GovRep   -> gov_report
%% QMSum    -> qmsum
%% MultiN   -> multi_news
%% TREC     -> trec
%% TriviaQA -> triviaqa
%% SAMSum   -> samsum
%% PCount   -> passage_count
%% PRE      -> passage_retrieval_en
%% Lcc      -> lcc
%% RB-P     -> repobench-p
%% ============================================================

\vspace{-3pt}
\subsection{RULER 16K}
\label{sec:experiments:ruler16k}

\begin{figure}[t]
  \centering
  \includegraphics[width=0.75\linewidth]{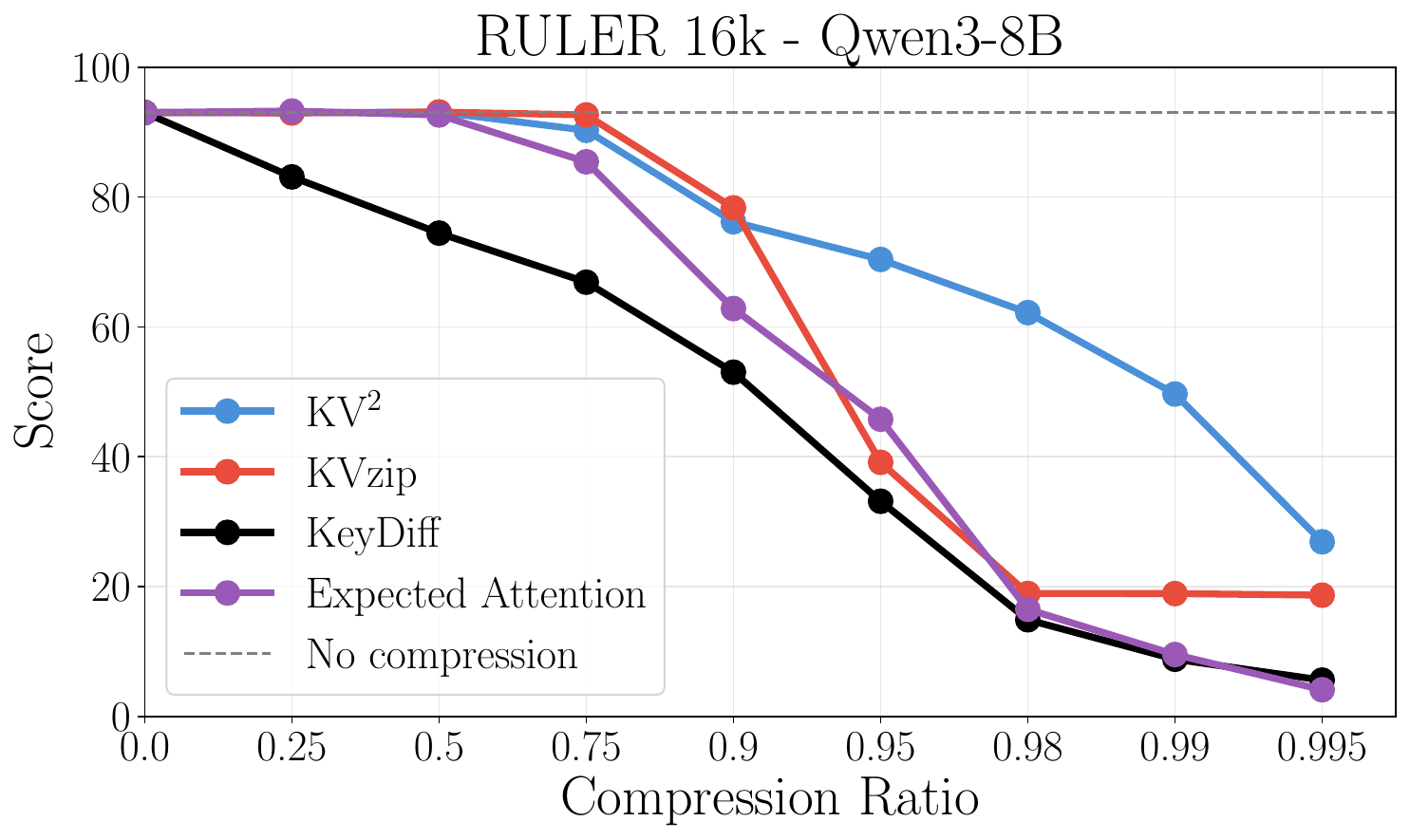}
  \caption[RULER 16K results on Qwen3-8B]{RULER 16K score versus compression ratio for \texttt{Qwen3-8B}. The largest gap appears in the severe-budget regime (compression ratio$>$90\%), where KV$^2$ remains much more robust than the compared baselines.}
  \label{fig:experiments:ruler16k}
  %\vspace{-10pt} % replaced by global float spacing
\end{figure}

RULER 16K is the clearest stress test for severe budgets, so we highlight the \texttt{Qwen3-8B} result in the main text (Figure~\ref{fig:experiments:ruler16k})
and defer the full per-backbone tables to Appendix~\ref{sec:appendix:ruler16k-llama} and Appendix~\ref{sec:appendix:ruler16k-qwen}. The same pattern holds on both backbones: at 5\% and especially 2\% KV-cache budgets, KV$^2$ degrades much more gracefully than the baselines. The advantage is strongest on sparse retrieval and aggregation, while distractor-heavy variants remain difficult for all methods. The gains include the multi-query and multi-value tasks, which require retrieving several needles from the same compressed cache (several keys for \texttt{mq}, all values of one key for \texttt{mv}): at 2\% on \texttt{Qwen3-8B}, KV$^2$ scores 81.45 and 90.60, while KVzip scores 0 on both (Table~\ref{tab:evaluation:ruler16k}). This is the regime where selective reconstruction matters most: with very little cache left, reconstructing a small set of informative anchors is more useful than spending capacity on less informative positions.\looseness=-1

\vspace{-3pt}
\subsection{Needle-in-a-Haystack (NIAH)}
\label{sec:experiments:needle}

\begin{figure}[t]
  \centering
  \includegraphics[width=0.8\linewidth]{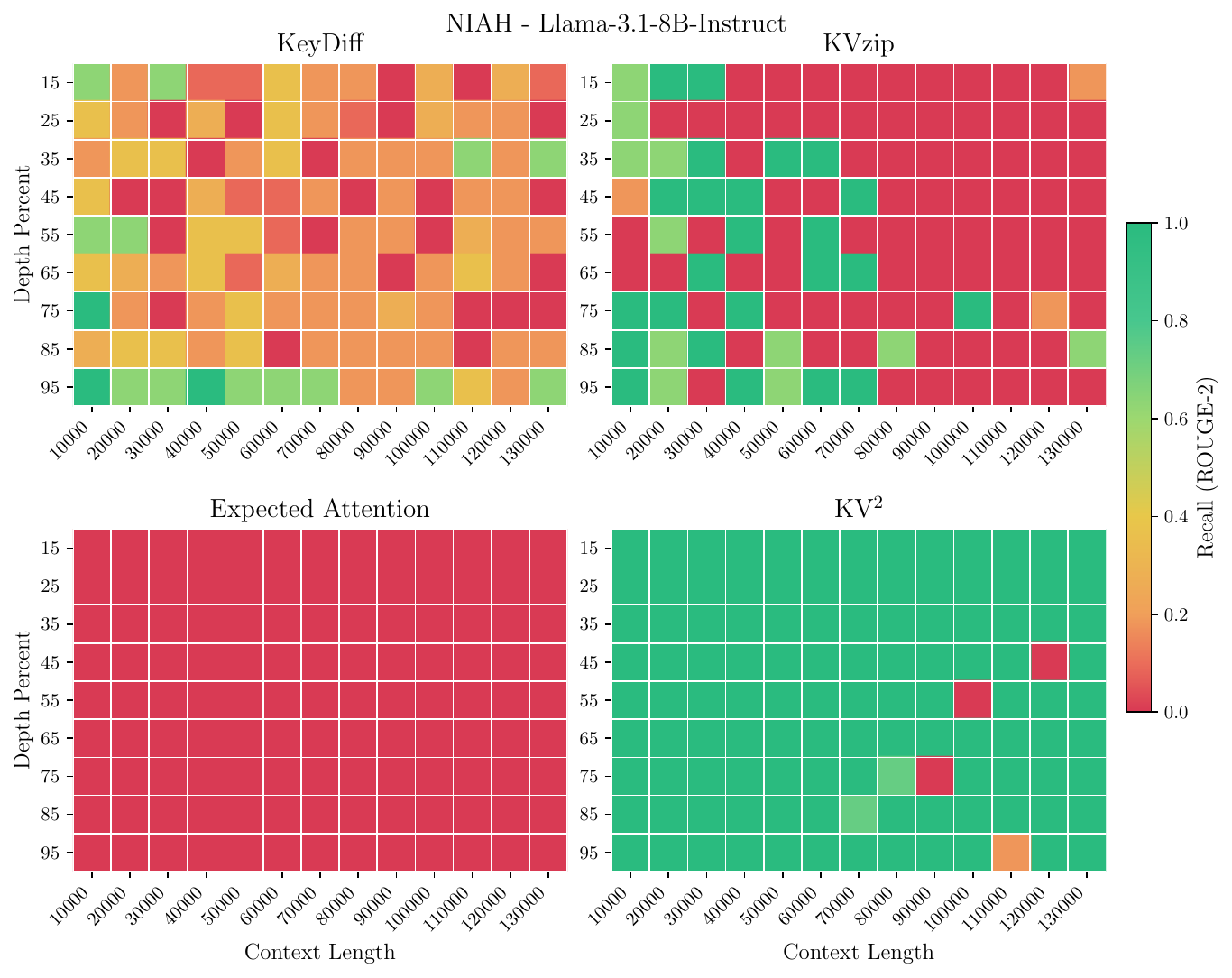}
  \caption[Needle-in-a-Haystack under compression]{Needle-in-a-Haystack results at a 5\% KV-cache budget on \texttt{Llama-3.1-8B-Instruct} across context lengths and needle depths.}
  \label{fig:experiments:needle}
\end{figure}

The NIAH experiment tests whether a target fact survives compression within a long distractor document. The distractor context is natural prose rather than RULER's repeated synthetic templates, providing a more realistic test case. Results are measured on \texttt{Llama-3.1-8B-Instruct} at a fixed 5\% KV-cache budget using per-example ROUGE-2 recall \cite{Hsieh2024,DevotoPaulGrahamEssays}. Figure~\ref{fig:experiments:needle} shows the same pattern as RULER in a simpler setting: KV$^2$ is the most stable method as context length and needle depth increase, whereas KeyDiff is patchy, and KVzip degrades more quickly on longer haystacks.\looseness=-1

\vspace{-3pt}
\subsection{LongBench}
\label{sec:experiments:longbench}

%Table~\ref{tab:evaluation:longbench} shows that the same pattern carries over to 
On LongBench (Table~\ref{tab:evaluation:longbench}),
a broader and more heterogeneous benchmark,
the same pattern carries over. The clearest gains again appear at tight budgets. At a 2\% KV-cache budget, KV$^2$ raises the average LongBench score from 24.16 to 33.59 on \texttt{Llama-3.1-8B-Instruct} and from 16.59 to 32.97 on \texttt{Qwen3-8B} relative to the next-best baseline. At 5\%, the gap remains large: KV$^2$ reaches 39.91 on Llama and 38.28 on Qwen, compared with 29.12 and 22.61 for the strongest competing methods. At 10\%, KV$^2$ still has the highest average (43.59 vs.\ 41.30 on Llama, 43.89 vs.\ 41.62 on Qwen), but this margin is small and we do not consider it conclusive. The gains are concentrated in multi-document question answering and in synthetic retrieval and counting tasks, where preserving a small set of supporting facts matters more than reconstructing context uniformly. The clearest recurring exception is the \texttt{RepoBench-P} subtask, where broader reconstruction remains more helpful. LongBench therefore sharpens the same boundary as the earlier diagnostics: selective reconstruction helps most when relevant evidence is sparse and anchor-like, and less when useful evidence is spread broadly across the prompt. Appendix~\ref{sec:appendix:infinitebench} extends this picture to longer inputs on a single 10\% subsample of InfiniteBench: KV$^2$ clearly leads at 2\% on \texttt{Llama-3.1-8B-Instruct} (38.06 vs.\ 33.50) and is on par with the strongest baseline (within 0.5 points) in all other settings.
% TODO (paJ1 W4): add an appendix listing the number of evaluated examples per sub-benchmark and reference it here.

\begin{figure*}[!ht]
  \centering
  \begin{minipage}[t]{0.40\linewidth}
    \centering
    \includegraphics[width=\linewidth]{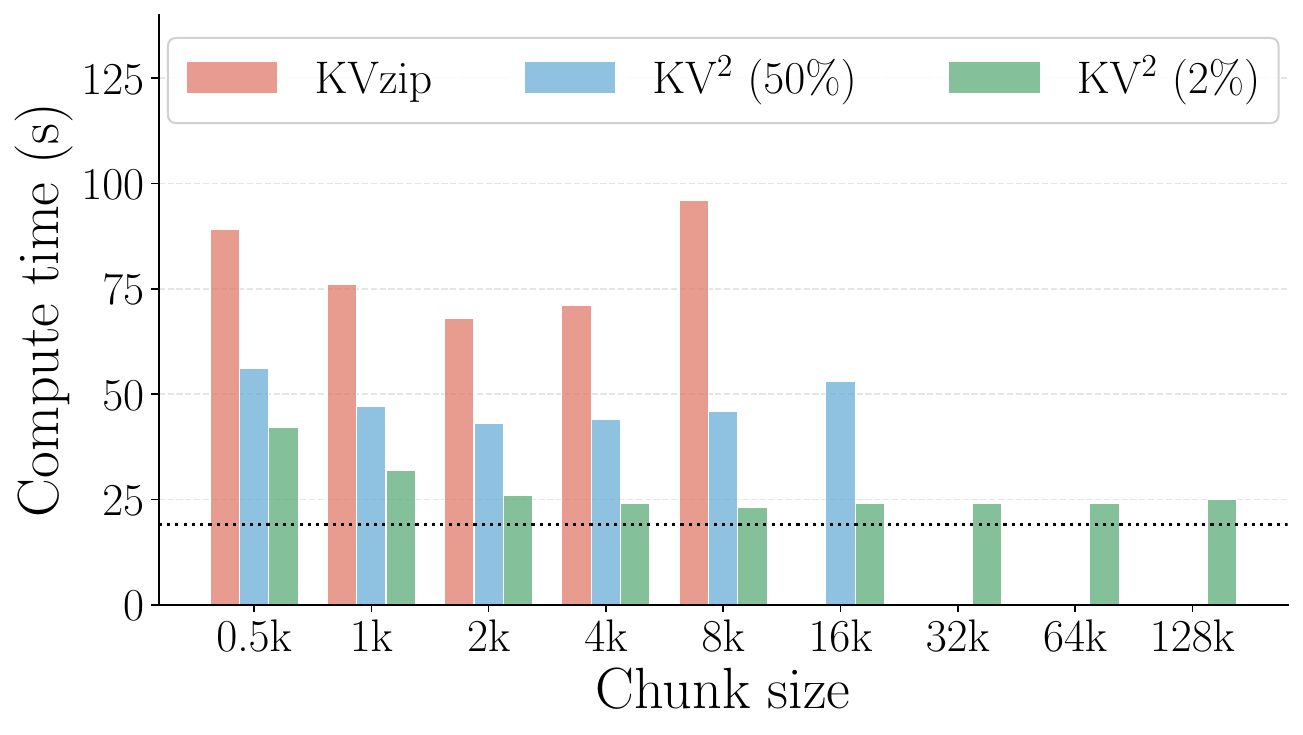}
  \end{minipage}\hfill
  \begin{minipage}[t]{0.40\linewidth}
    \centering
    \includegraphics[width=\linewidth]{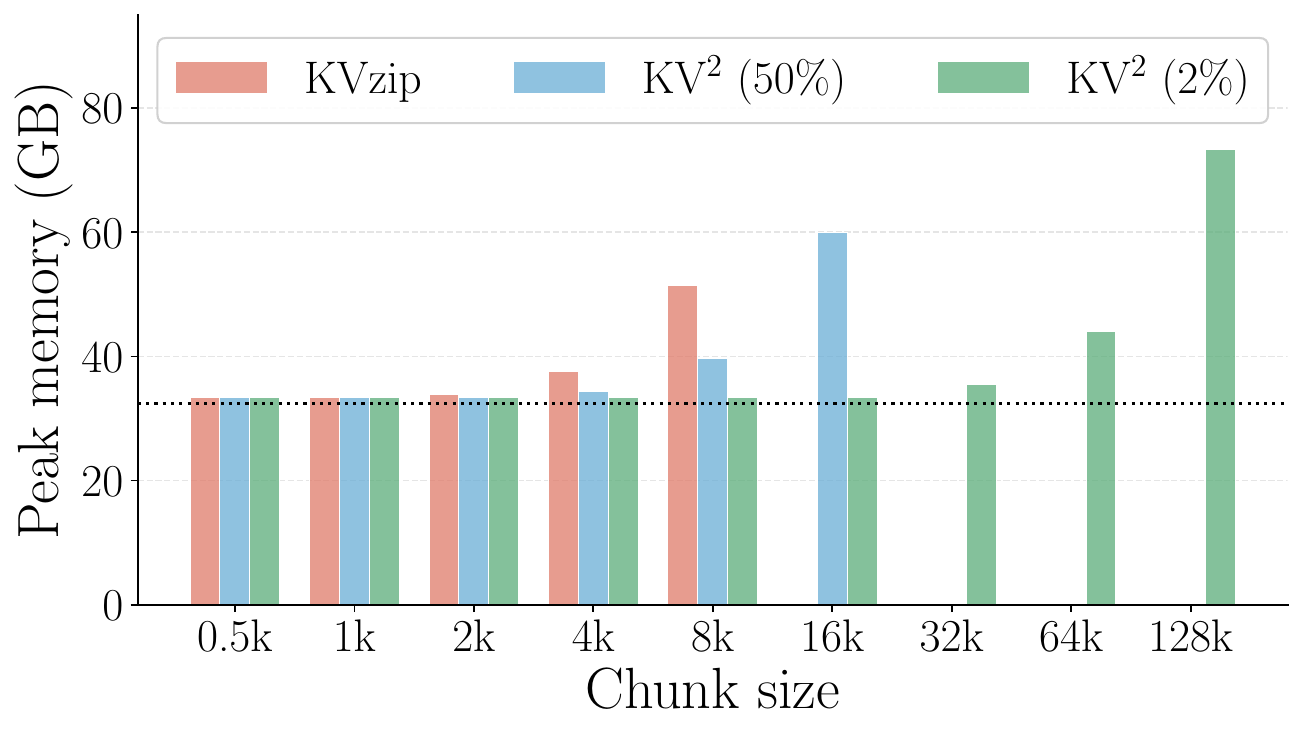}
  \end{minipage}
  \caption[Single-pass runtime and memory versus chunk size]{Compression-stage runtime (left) and peak memory (right) versus chunk size for single-pass KV$^2$ compared with KVzip. For KV$^2$, the value in parentheses in the legend denotes the reconstruction percentage retained from the proxy-selected query set in the first stage. Missing bars indicate out-of-memory configurations.}
  \label{fig:analysis:single-pass}
  %\vspace{-10pt} % replaced by global float spacing
\end{figure*}

\vspace{-3pt}
\section{Analysis}
\label{sec:analysis}

The benchmark results above highlight the empirical strengths of KV$^2$. This section examines how the method shapes this behavior: how much compression-stage runtime and memory it saves relative to KVzip (\S\ref{sec:analysis:single-pass}), how the first-stage proxy and reconstruction budget shape the reconstruction set (\S\ref{sec:analysis:proxy}, \S\ref{sec:analysis:replay-budget}), and how much additional quality iterative refinement can recover once those pieces are fixed (\S\ref{sec:analysis:iterative-refinement}). These efficiency measurements concern only the compression stage. In the current \texttt{kvpress} backend, both KVzip and KV$^2$ are realized through headwise masking rather than physically compacted caches, so the reported runtime and peak memory should be read as compression-stage costs
%do not yet represent 
instead of realized decode-time savings from a production serving backend. Unless stated otherwise, this section uses \texttt{Llama-3.1-8B-Instruct} with 124K context tokens on an NVIDIA H100 GPU in FP16 with FlashAttention-2 \cite{Dao2023}.\looseness=-1

\vspace{-3pt}
\subsection{Single-pass efficiency}
\label{sec:analysis:single-pass}

The clearest practical advantage of KV$^2$ over KVzip is single-pass efficiency. Figure~\ref{fig:analysis:single-pass} shows that, within this masking-based backend, it is consistently faster than KVzip and uses less peak memory across the relevant chunk sizes. The reason is structural rather than incidental: KV$^2$ reconstructs only a proxy-selected subset of queries, so the compression stage performs less work and keeps a smaller active reconstruction footprint. The chunk-size curves in Figure~\ref{fig:analysis:single-pass} also show why 4K is the default chunk size.
Below 4K, the repeated per-chunk overhead drives compression-stage runtime up, as shown in the left subfigure.
Above 4K, the reconstruction set grows linearly with chunk size, raising both runtime and peak memory until the largest configurations run out of memory (missing bars in the figure).
%The chunk-size curves also explain the 4K default: smaller chunks pay repeated per-chunk overhead, while larger chunks increase the size of the reconstruction set. 
In contrast, the quality-side picture is comparatively flat: across length buckets in LongBench-E \cite{Bai2023} (LongBench with a more even length distribution), KV\textsuperscript{2}'s score is largely insensitive to chunk size, with larger chunks helping mainly on the longest inputs (Appendix~\ref{sec:appendix:longbench-e}). 
%Appendix~\ref{sec:appendix:longbench-e} provides the complementary quality-side view: across length buckets in LongBench-E \cite{Bai2023} (LongBench with a more even length distribution), KV$^2$ stays relatively stable over a broad chunk-size range, with larger chunks helping mainly on the longest inputs. 
A chunk size of 4K therefore captures nearly all the available quality while remaining highly efficient.
To separate nominal from physical savings: for \texttt{Llama-3.1-8B-Instruct} with 124K tokens in FP16, the full cache occupies 15.14\,GiB. At a 2\% budget, a physically compacted cache would occupy 0.303\,GiB, whereas the masking backend keeps the full 15.14\,GiB resident and adds 0.695\,GiB of mask metadata (3.8\,MiB if stored as a bitmask). The compression stage additionally allocates a 60.5\,MiB score tensor and transient score-extraction buffers of 20.7\,MiB for KV$^2$ versus 512.5\,MiB for KVzip.
%Taken together, the efficiency curves and LongBench-E make 4K the best balanced operating point.

\vspace{-3pt}
\subsection{Proxy scorer}
\label{sec:analysis:proxy}

\begin{figure}[t]
  \centering
  \includegraphics[width=0.75\linewidth]{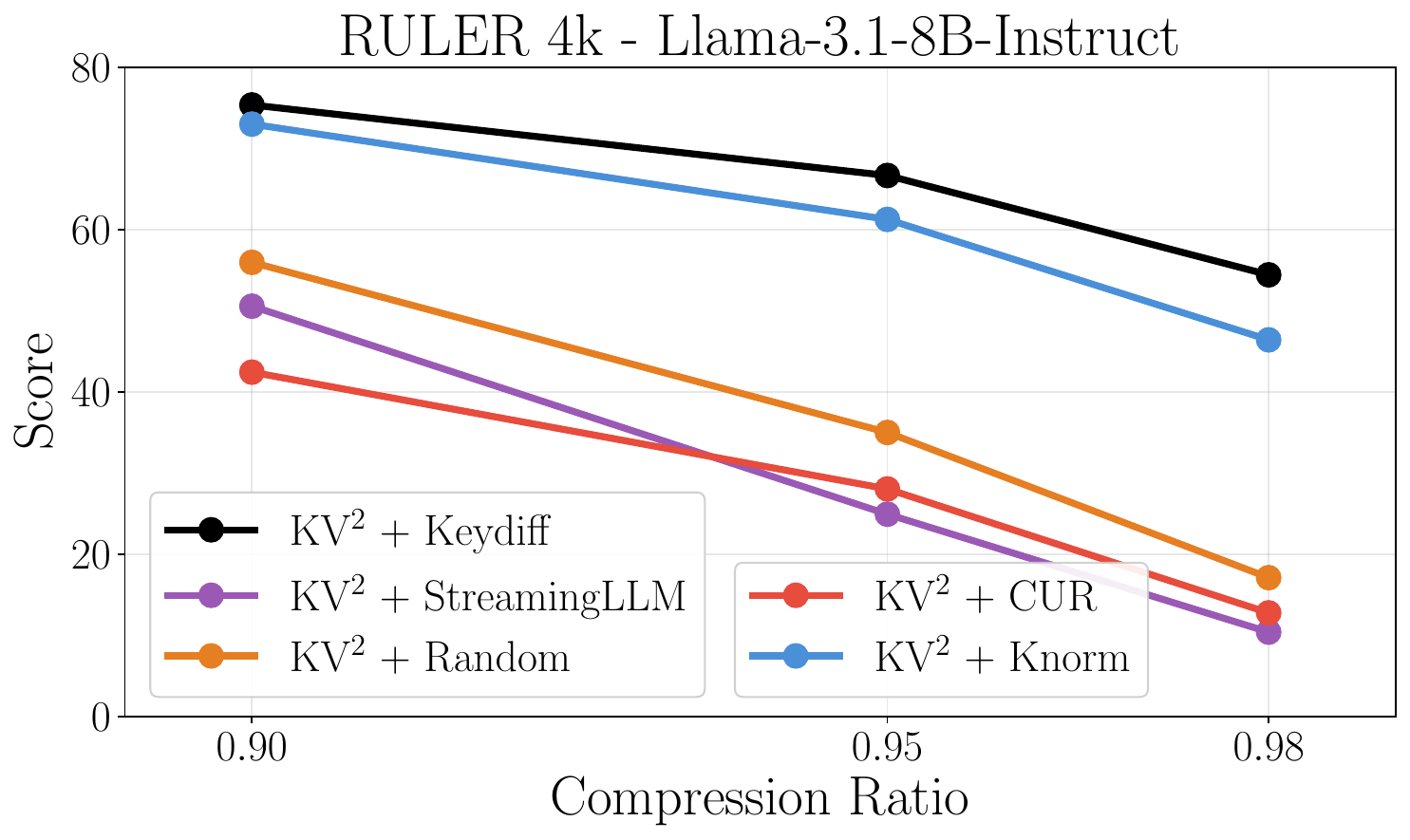}
  \caption[Accuracy versus proxy scorer]{RULER 4K score versus compression ratio for \texttt{Llama-3.1-8B-Instruct} when KV$^2$ is paired with different first-stage proxy scorers.}
   %\vspace{-10pt} % replaced by global float spacing
  \label{fig:analysis:proxy-scorer}
\end{figure}

\begin{figure*}[!ht]
  \centering
  \begin{minipage}[t]{0.40\linewidth}
    \centering
    \includegraphics[width=\linewidth]{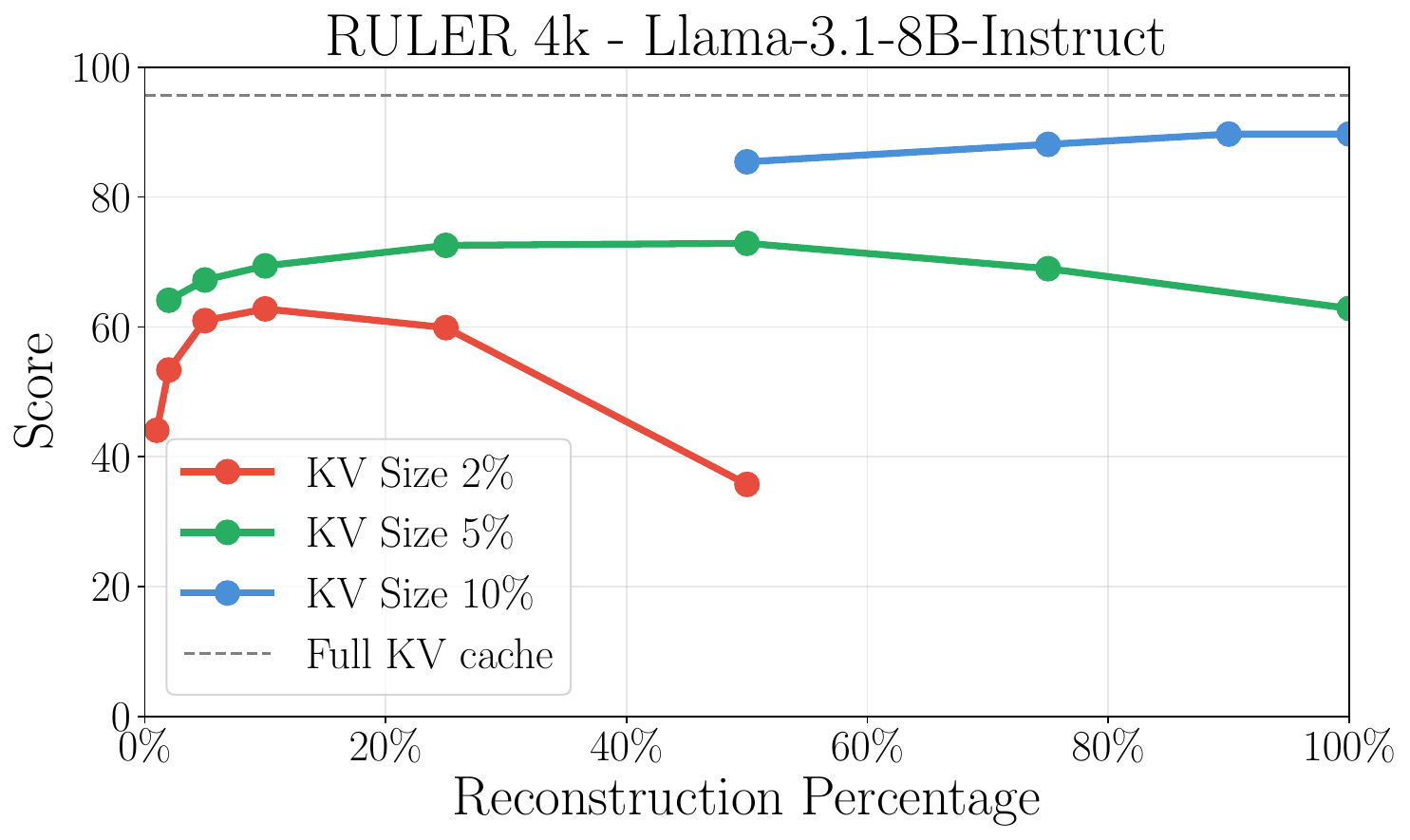}
  \end{minipage}\hfill
  \begin{minipage}[t]{0.40\linewidth}
    \centering
    \includegraphics[width=\linewidth]{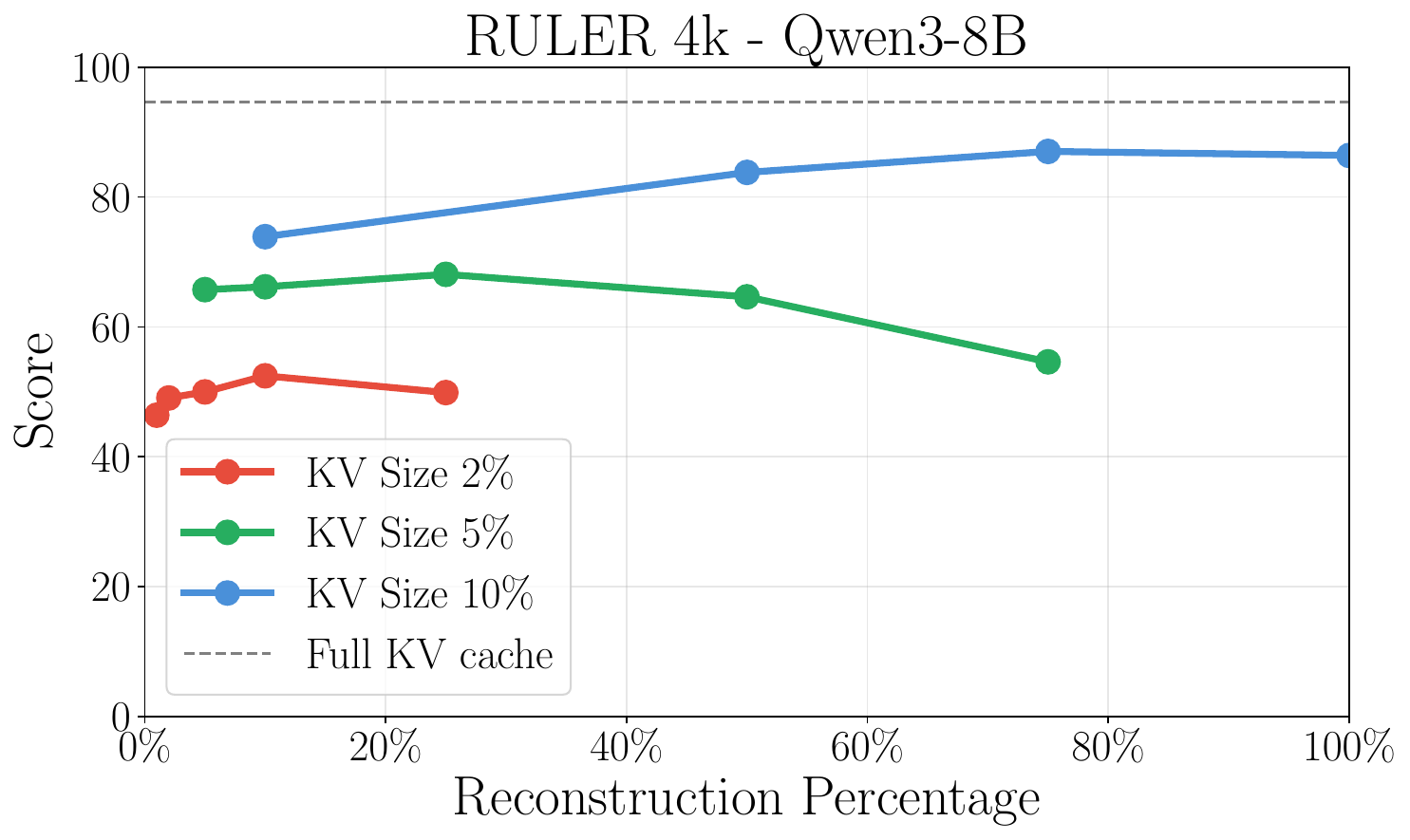}
  \end{minipage}
    \caption[Accuracy versus reconstruction percentage]{RULER 4K score versus reconstruction percentage for \texttt{Llama-3.1-8B-Instruct} (left) and \texttt{Qwen3-8B} (right). In the legend, \texttt{KV Size} denotes the final retained KV-cache budget.}
  %\vspace{-10pt}
  \label{fig:analysis:reconstruction}
\end{figure*}

The first-stage proxy mechanism matters, even though it is not the final scoring rule. Figure~\ref{fig:analysis:proxy-scorer} isolates this effect. All curves are KV$^2$ with the same second stage, and only the proxy scorer changes. KeyDiff \cite{Park2025} is the strongest option throughout, with Knorm \cite{Devoto2024} as the closest alternative. CUR \cite{Sengupta2025} and especially StreamingLLM \cite{Xiao2023} fall off much more quickly as compression tightens, 
with both dropping below a random-proxy baseline at the highest compression ratios studied. 
%even below the random proxy baseline at the high compression ratios studied. 
A plausible reason is that StreamingLLM is designed to preserve sink tokens and recent context for streaming generation, and both StreamingLLM and CUR rely on structurally fixed selection patterns rather than content-adaptive ones.
Therefore, under severe compression, a content-adaptive proxy set can cover the useful signal better. The random proxy also acts as a control for the replay design itself: it replays the same number of tokens at the same appended positions, yet stays far below KeyDiff, and StreamingLLM, whose proxy prioritizes sink and recent positions, ends lowest at 98\% compression. The gain therefore comes from which tokens are replayed rather than from recency or duplication alone. Appendix~\ref{sec:appendix:selection-example} shows the same effect qualitatively for the default KeyDiff proxy after layer- and head-averaging: the proxy mass concentrates on the instruction, assignment chain, and answer span, while only a sparse set of tokens is drawn from the repetitive filler context. The proxy is thus swappable but not inconsequential: the gain from reconstruction scoring is separable from the proxy (\S\ref{sec:analysis:iterative-refinement}), but the proxy determines how much of it is realized, especially under severe compression, and KeyDiff is our recommended default.

% Defined early so that it can float to the top of a column on the same page as §5.4.
\begin{figure}[t]
  \centering
  \includegraphics[width=0.8\linewidth]{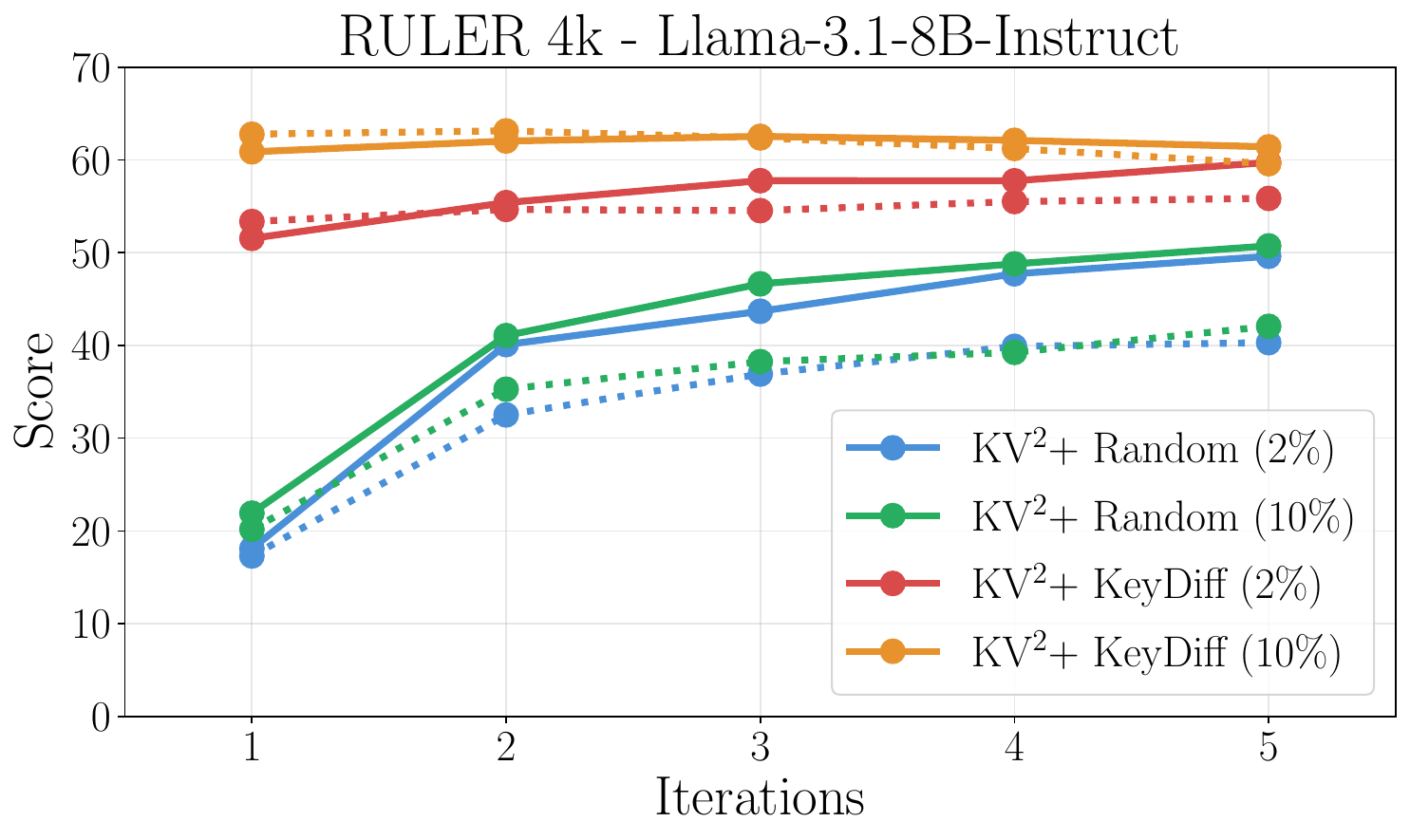}
  \caption[Iterative refinement on RULER 4K]{RULER 4K score over refinement iterations at a 2\% KV-cache budget for \texttt{Llama-3.1-8B-Instruct}. Iteration~1 is single-pass KV$^2$. Solid lines use KVzip+ normalization \cite{Jegou2026}, while dotted lines use the unnormalized scores of Eq.~\ref{eq:method:kvsquared} as in all other experiments. The legend lists the first-stage proxy scorer and the reconstruction percentage. Most gains appear in the first few passes, after which improvements diminish.}
  \label{fig:analysis:ruler-iterative}
\end{figure}

\vspace{-3pt}
\subsection{Reconstruction budget}
\label{sec:analysis:replay-budget}

Figure~\ref{fig:analysis:reconstruction} makes the reconstruction-budget trade-off explicit. 
On both backbones, the optimal reconstruction fraction grows with the final KV-cache budget. 
For \texttt{Llama-3.1-8B-Instruct}, the RULER score peaks at 10\%, 50\%, and 100\% reconstruction for final budgets of 2\%, 5\%, and 10\%, respectively.
For \texttt{Qwen3-8B}, the corresponding maxima are 10\%, 25\%, and 75\%. 
These trends demonstrate that more retained cache rewards more reconstruction, and no single $\rho$
is optimal across budgets.
A plausible explanation is that under severe budgets, the retained cache consists of sparse anchors, so what matters is ranking a few candidates correctly, and a small, well-chosen query set concentrates the reconstruction signal on them. As the budget grows, the retained set must also cover distributed, non-anchor evidence that only queries attending to it can identify, so broader reconstruction pays off and the optimum moves toward full replay.
This monotonic relationship strengthens the default schedule in Section~\ref{sec:experiments:setup}, which sets $\rho$ to track the final retained KV-cache ratio,
where tighter budgets get smaller reconstruction sets, looser budgets get larger ones.
While this matched default does not reach the per-budget optimum, it does capture the right scaling and keeps reconstruction cost proportional to the budget.
%On both backbones, the optimal reconstruction fraction increases with %moves upward as 
%the final KV budget, 
%but it does so in a structured way rather than requiring broad reconstruction by default. For \texttt{Llama-3.1-8B-Instruct}, the RULER score peaks at 10\%, 50\%, and 100\% reconstruction for final KV sizes of 2\%, 5\%, and 10\%, respectively; for \texttt{Qwen3-8B}, the corresponding maxima occur at 10\%, 25\%, and 75\%. 
%Tight budgets thus peak with very small reconstruction sets, %The main lesson is that tight budgets peak very early, 
%while looser budgets continue to benefit from broader reconstruction before leveling off. 
%This pattern also motivates the default schedule described in Section \ref{sec:experiments:setup}: the reconstruction fraction simply matches the final retained KV-cache ratio, so tighter budgets use smaller reconstruction sets, while larger retained budgets justify broader reconstruction.
%This is also why the default schedule makes the replay fraction grow with the amount of KV cache that is retained: as compression becomes more aggressive, the useful replay fraction shrinks, whereas larger retained KV budgets make broader replay increasingly worthwhile.

\vspace{-3pt}
\subsection{Iterative refinement}
\label{sec:analysis:iterative-refinement}

The proxy-scorer and reconstruction-budget results suggest a next question: once one KV$^2$ pass has produced a useful but potentially imperfect reconstruction set, how much can later passes improve it? Figure~\ref{fig:analysis:ruler-iterative} answers this on RULER 4K at a 2\% KV-cache budget, both with the default unnormalized scores of Eq.~\ref{eq:method:kvsquared} and with KVzip+ normalization \cite{Jegou2026}, which rescales each attention weight by the norm of the cached value's contribution to the attention output, $\lVert W_O v_i \rVert$, divided by the norm of the reconstruction token's hidden state, $\lVert h_j \rVert$.
KeyDiff plays a different role here than in the baseline comparisons: as a baseline it is the eviction rule, whereas inside KV$^2$ it only proposes reconstruction queries and eviction follows the Stage~2 scores. Most of the gain comes from this change of eviction rule and is already contained in the first iteration: KeyDiff alone scores 23.22 (Appendix~\ref{sec:appendix:ruler4k}), whereas single-pass KV$^2$ with the KeyDiff proxy reaches 53.36.
Later iterations help most when the initial proxy is weak. With a random proxy, single-pass KV$^2$ scores only 17-22, and five iterations raise this to 40-42 without normalization and to about 50 with it. With the KeyDiff proxy, the gain is smaller (53.4 to 55.9 without normalization, 51.6 to 59.7 with it), and at a 10\% reconstruction fraction, refinement brings no consistent gain.
KVzip+ normalization thus slightly lowers single-pass quality with the KeyDiff proxy but makes refinement more effective in every setting.
%such as with a less effective proxy or a very small reconstruction fraction; once the first pass is already strong, later passes add little.
Iterative refinement is therefore most attractive as a targeted extension for harder settings rather than as the default configuration.
Moreover, Appendix~\ref{sec:appendix:iterative-refinement-cost} shows that these gains come mainly at a runtime cost, while peak memory changes much less.\looseness=-1

\paragraph{Summary}
The analyses in this section identified the main operating regime of our method. The single-pass version is the strongest practical default because it already delivers the main quality gains while staying clearly cheaper than broader reconstruction (\S\ref{sec:analysis:single-pass}). 
The proxy scorer is swappable but not inconsequential. Its quality matters most at severe compression, and KeyDiff is our recommended default (\S\ref{sec:analysis:proxy}).
Reconstruction percentage provides a useful deployment knob (\S\ref{sec:analysis:replay-budget}), while iterative refinement can recover additional quality mainly when the initial proxy is weak (\S\ref{sec:analysis:iterative-refinement}).

\vspace{-3pt}
\section{Conclusion} \label{sec:discussion}
In this work, we showed that reusable KV-cache compression does not need to choose between cheap heuristics and reconstructing everything. KV\textsuperscript{2} shows that a small, well-chosen reconstruction set can already preserve much of the useful long-context signal under severe compression.
The two-stage structure, paired with the observation that informative anchors are sparse at extreme budgets, points toward compression methods that are both cheaper and more accurate.
Our efficiency results cover the compression stage under a masking backend. Realizing the corresponding memory and decoding savings in paged serving systems is the main remaining step (see Limitations).

% \paragraph{Practical relevance.}
% KV$^2$ is particularly well matched to deployments with long shared prefills and tight per-context memory pressure: multi-turn assistants that repeatedly revisit conversation history \cite{Gao2024}, long-document question answering systems that issue many short follow-up queries against the same source context \cite{Srivatsa2024}, and tool- or agent-style workloads that pause and resume around a large shared prompt prefix \cite{Srivatsa2024,Abhyankar2024}. What these settings share is that prompt length dominates generation length, reusable caches compete for limited high-bandwidth memory, and each later request often depends on recovering a few decisive pieces of evidence rather than the full prompt in uniform detail. 
% Selective reconstruction is well-aligned with this pattern, by spending its compression budget on identifying the small set of positions that carry the relevant signal, rather than uniformly summarizing the prompt.
% Crucially, this robustness does not require an impractically expensive compression procedure. Within the current masking-based \texttt{kvpress} backend, the single-pass variant already provides the best overall quality-efficiency balance, because it reconstructs only a selected subset of queries and is consistently cheaper than KVzip in both compression-stage runtime and peak memory.

\section*{Limitations}
Three directions appear most important to improve the current work. 

First, iterative refinement should be explored further and should become adaptive rather than fixed. Section~\ref{sec:analysis:iterative-refinement} shows that additional passes help mainly when the initial proxy is weak or the reconstruction budget is very small, while Appendix~\ref{sec:appendix:iterative-refinement-cost} shows that these gains come mainly at a runtime cost. This suggests refinement schedules that continue only while the reconstruction set or the resulting scores are still changing meaningfully, and perhaps allocate more compression-time compute only to harder retrieval or multi-document workloads. 

Second, the first-stage proxy mechanism itself should be strengthened. Because KV\textsuperscript{2} is a modular two-stage framework, better proxy scorers should improve both single-pass quality and the value of later refinement. 
The proxy ablation in Section~\ref{sec:analysis:proxy} suggests that content-adaptive scorers outperform structurally fixed ones at low budgets, suggesting that learned or model-aware proxies may be able to improve further.

Third, the current masking-based realization should be replaced by a compact serving backend. Within the \texttt{kvpress} toolkit, KV\textsuperscript{2} measures compression-stage runtime and memory faithfully, but it does not yet materialize physically compact reusable caches due to the toolkit's implementations. Headwise cache compaction, optimized attention kernels that allow each head's individually compressed sequence length, and repeated-query serving pipelines are therefore needed to translate the present compression results into full decoding-time memory and latency gains.

\section*{Acknowledgment}
We thank the reviewers for helpful comments.
Part of this work was funded by the KiKIT (The Pilot
Program for Core-Informatics at the KIT) of the
Helmholtz Association.
Part of this work was supported by European Union's Horizon Europe programme grant agreement No.~101213369
(DVPS).
The authors gratefully acknowledge the computing time provided on the high-performance computer HoreKa by the National High-Performance Computing Center at KIT (NHR@KIT). This center is jointly supported by the Federal Ministry of Research, Technology and Space and the Ministry of Science, Research and the Arts of Baden-Württemberg, as part of the National High-Performance Computing (NHR) joint funding program. HoreKa is partly funded by the German Research Foundation (DFG).

\bibliography{references}

%\clearpage
\appendix
%\raggedbottom

\section{Complexity Analysis}
\label{sec:appendix:complexity}
Let $n_c$ denote the prefilled context length. The complexity is easiest to understand by separating KV\textsuperscript{2} into the lightweight proxy stage and the two computations inside selective reconstruction. The proxy stage is inexpensive. In the default implementation, the efficient KeyDiff proxy computes one anchor-style key score per token, so proxy scoring is linear in context length, $O(n_c)$.

The reconstruction forward pass is the expensive part. For one chunk, the reconstruction block has length $n_q \approx \rho m$. Because these reconstruction queries are appended after the prefilled document, each one attends over the full cached context and also over earlier reconstruction tokens in the same block. The resulting per-chunk cost is therefore $O(n_c n_q + n_q^2/2)$. Repeating this across the roughly $n_c/m$ chunks gives a total reconstruction-forward cost of $O(\rho n_c^2 + \rho^2 n_c m / 2)$. However, this reconstruction forward remains an ordinary causal pass over the shared cache, so it can still use optimized attention implementations such as FlashAttention-style kernels during the forward computation \cite{Dao2023}.

Score extraction is cheaper because it is chunk-local. After the reconstruction forward pass, KV\textsuperscript{2} recomputes attention only on a reduced matrix of size $n_q \times (m+n_q)$ for the current chunk, as illustrated in Figure~\ref{fig:method:attention-matrix}. This gives $O(n_q(m+n_q)) = O(\rho m^2)$ work per chunk and $O(\rho m n_c)$ over all chunks. Unlike the reconstruction forward, however, the final score extraction requires a Softmax over keys followed by a max over queries, a cross-dimensional dependency that does not map cleanly onto the usual blockwise attention algorithms \cite{Kim2025}.

\section{Iterative refinement cost}
\label{sec:appendix:iterative-refinement-cost}

\begin{figure*}[ht!]
  \centering
  \begin{minipage}[t]{0.49\linewidth}
    \centering
    \includegraphics[width=\linewidth]{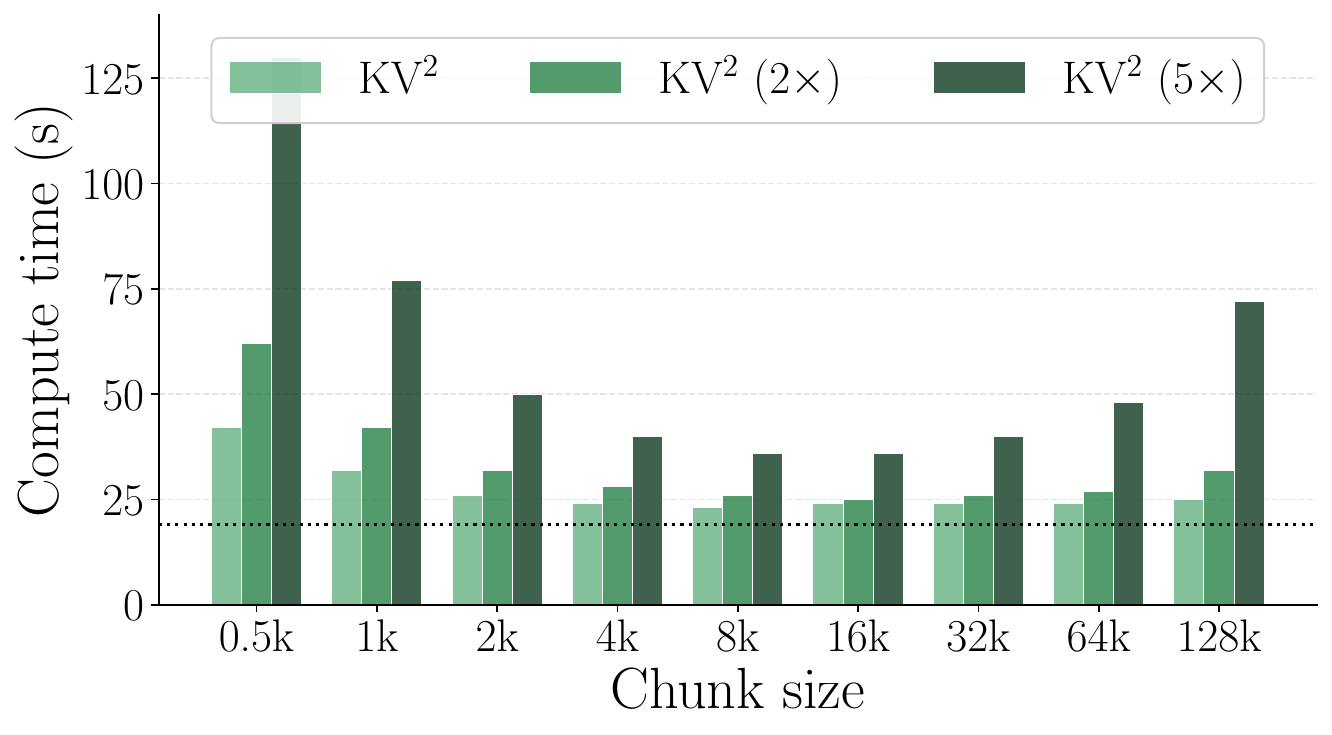}
  \end{minipage}\hfill
  \begin{minipage}[t]{0.49\linewidth}
    \centering
    \includegraphics[width=\linewidth]{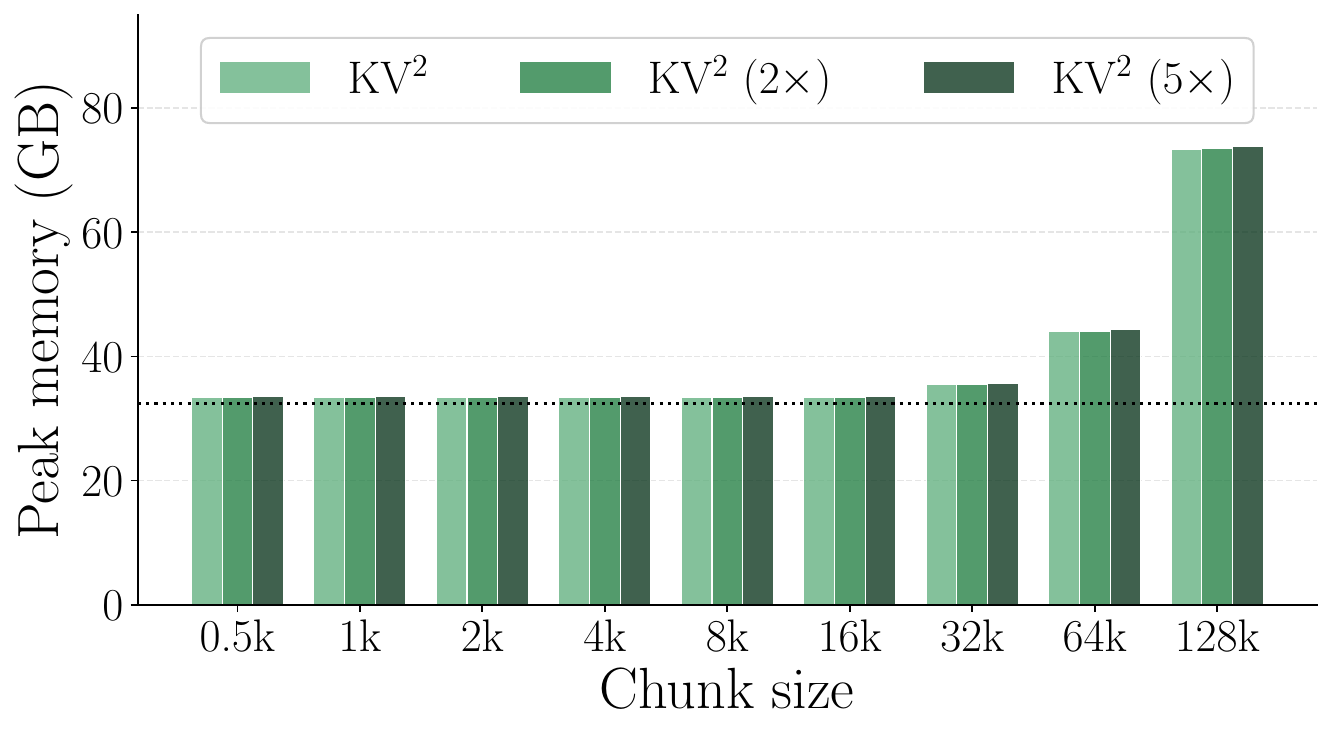}
  \end{minipage}
  \caption[Iterative runtime and memory versus chunk size]{Compression-stage runtime (left) and peak memory (right) versus chunk size for iterative KV$^2$. Additional refinement passes increase runtime roughly in proportion to the number of passes, while peak memory is driven mainly by chunk size and changes much less with the number of iterations. All curves use a fixed 2\% reconstruction percentage. The number in parentheses in the legend denotes the number of refinement iterations.}
  \label{fig:analysis:iterative-cost}
\end{figure*}

Section~\ref{sec:analysis:iterative-refinement} showed that iterative refinement can improve quality, especially when the initial proxy is weak or the reconstruction fraction is small. The remaining question is cost. Figure~\ref{fig:analysis:iterative-cost} shows that runtime rises roughly in proportion to the number of refinement passes, especially when chunk size is small and many local reconstructions must be repeated. Peak memory remains driven mainly by chunk size because each pass reuses essentially the same working buffer. This is why iterative refinement is best viewed as an optional extension for harder settings rather than the default configuration.

\section{LongBench-E}
\label{sec:appendix:longbench-e}

LongBench-E complements Section~\ref{sec:analysis:single-pass} by separating input-length effects within the same task families at a fixed 5\% KV budget. Table~\ref{tab:evaluation:longbench-e} reports the results across these length buckets. All three KV$^2$ variants remain ahead of KeyDiff and KVzip on the overall average, and they stay relatively close to one another. Together with Figure~\ref{fig:analysis:single-pass}, this supports the 4K default used in the main paper.

%% LongBench-E: QA/summarization and classification/counting/code (single float, one caption)
\begin{table*}[!htbp]
\centering
\thesistablestyle
\caption[LongBench-E by input length]{LongBench-E results for \texttt{Llama-3.1-8B-Instruct} at 5\% KV retention, grouped by task and input length. The table columns labeled 4k, 8k, 16k, and 16k+ denote LongBench-E input-length buckets measured in words, whereas the method labels KV² (2K), KV² (8K), and KV² (32K) denote chunk sizes measured in tokens. These are therefore different quantities and should not be read on the same scale. Across both suites, KV² consistently outperforms KeyDiff and KVzip on average, and larger chunk sizes tend to be strongest on longer inputs while smaller chunks remain more competitive on shorter inputs.}
\label{tab:evaluation:longbench-e}

\begin{adjustbox}{max totalsize={\textwidth}{0.34\textheight},center}
\begin{tabular*}{\textwidth}{@{\extracolsep{\fill}}lccccccccccccc@{}}
\toprule
\multirow{3}{*}{\textbf{Method}}
    & \multicolumn{4}{c}{\textbf{Single-doc QA}}
    & \multicolumn{4}{c}{\textbf{Multi-doc QA}}
    & \multicolumn{5}{c}{\textbf{Summ.}} \\
\cmidrule(lr){2-5} \cmidrule(lr){6-9} \cmidrule(lr){10-14}
    & \multicolumn{4}{c}{NrtvQA}
    & \multicolumn{4}{c}{Hotpot}
    & \multicolumn{5}{c}{QMSum} \\
\cmidrule(lr){2-5} \cmidrule(lr){6-9} \cmidrule(lr){10-14}
    & 8k & 16k & 16k+ & $\mu$
    & 4k & 8k & 16k & $\mu$
    & 4k & 8k & 16k & 16k+ & $\mu$ \\
\midrule

Full
    & 35.06 & 21.48 & 34.12 & 30.85
    & 63.77 & 59.47 & 59.45 & 60.9
    & 22.14 & 26.34 & 25.32 & 23.54 & 25.21 \\

% \midrule
% \multicolumn{14}{c}{\textit{KV Size = 2\%}} \\
% \midrule
%
% KeyDiff
%     & -- & -- & -- & --
%     & -- & -- & -- & --
%     & -- & -- & -- & -- & -- \\
%
% KVzip
%     & -- & -- & -- & --
%     & -- & -- & -- & --
%     & -- & -- & -- & -- & -- \\
%
% KVsquared(KeyDiff 2\%)
%     & -- & -- & -- & --
%     & -- & -- & -- & --
%     & -- & -- & -- & -- & -- \\

\midrule
\multicolumn{14}{c}{\textit{KV Size = 5\%}} \\
\midrule

KeyDiff
    & 19.87 & 14.12 & 28.51 & 23.29
    & 26.27 & 24.45 & 30.93 & 27.22
    & 15.58 & 19.34 & 21.0 & 18.69 & 19.97 \\

KVzip
    & 19.9 & 15.07 & 27.32 & 22.86
    & 36.18 & 27.28 & 33.44 & 32.3
    & \textbf{20.08} & 21.62 & 21.24 & \textbf{21.58} & 21.31 \\

KV² (2K)
    & \textbf{23.87} & 19.59 & 34.51 & 28.83
    & \textbf{48.68} & \textbf{53.58} & 45.06 & 49.1
    & \textbf{20.08} & 21.62 & 21.24 & \textbf{21.58} & 21.31 \\

KV² (8K)
    & 20.7 & 15.96 & 32.71 & 26.33
    & 45.13 & 51.12 & \textbf{53.95} & \textbf{50.07}
    & 20.07 & \textbf{24.81} & 23.52 & 19.11 & \textbf{23.19} \\

KV² (32K)
    & 19.24 & \textbf{22.36} & \textbf{38.0} & \textbf{30.87}
    & 45.13 & 49.41 & 53.18 & 49.24
    & 20.07 & 24.22 & \textbf{23.59} & 19.73 & 23.13 \\

\bottomrule
\end{tabular*}
\end{adjustbox}

\medskip

\begin{adjustbox}{max totalsize={\textwidth}{0.34\textheight},center}
\begin{tabular*}{\textwidth}{@{\extracolsep{\fill}}lccccccccccccccc@{}}
\toprule
\multirow{3}{*}{\textbf{Method}}
    & \multicolumn{4}{c}{\textbf{Few-shot / Cls.}}
    & \multicolumn{4}{c}{\textbf{Synth. Count}}
    & \multicolumn{5}{c}{\textbf{Code}}
    & \multirow{3}{*}{\textbf{Avg.}} \\
\cmidrule(lr){2-5} \cmidrule(lr){6-9} \cmidrule(lr){10-14}
    & \multicolumn{4}{c}{TREC}
    & \multicolumn{4}{c}{PCount}
    & \multicolumn{5}{c}{LCC}
\\
\cmidrule(lr){2-5} \cmidrule(lr){6-9} \cmidrule(lr){10-14}
    & 4k & 8k & 16k & $\mu$
    & 4k & 8k & 16k & $\mu$
    & 4k & 8k & 16k & 16k+ & $\mu$
\\
\midrule

Full
    & 30.0 & 22.0 & 14.0 & 22.0
    & 28.0 & 13.3 & 19.0 & 20.1
    & 49.06 & 43.97 & 38.36 & 60.00 & 44.16
    % & 48.04 & 39.37 & 38.88 & 47.0 & 42.2
    & 33.87 \\

% \midrule
% \multicolumn{19}{c}{\textit{KV Size = 2\%}} \\
% \midrule
%
% KeyDiff
%     & -- & -- & -- & --
%     & -- & -- & -- & --
%     & -- & -- & -- & -- & --
%     & -- & -- & -- & -- & -- \\
%
% KVzip
%     & -- & -- & -- & --
%     & -- & -- & -- & --
%     & -- & -- & -- & -- & --
%     & -- & -- & -- & -- & -- \\
%
% KVsquared(KeyDiff 2\%)
%     & -- & -- & -- & --
%     & -- & -- & -- & --
%     & -- & -- & -- & -- & --
%     & -- & -- & -- & -- & -- \\

\midrule
\multicolumn{15}{c}{\textit{KV Size = 5\%}} \\
\midrule

KeyDiff
    & 23.0 & 35.0 & 25.0 & 27.67
    & 14.0 & 10.0 & 15.0 & 13.0
    & 16.53 & 21.91 & 18.41 & 24.2 & 19.05
    % & 51.39 & 40.83 & 37.92 & 30.5 & 43.28
    & 21.70 \\

KVzip
    & 7.0 & 13.0 & 17.0 & 12.33
    & 17.0 & 5.0 & 6.0 & 9.33
    & 30.52 & 25.56 & 19.09 & 23.67 & 25.06
    % & 53.22 & 49.65 & 49.7 & 22.33 & 50.45
    & 20.53 \\

KV² (2K)
    & 29.5 & 41.0 & 38.0 & 36.17
    & \textbf{28.55} & \textbf{15.17} & \textbf{18.0} & \textbf{20.57}
    & \textbf{45.14} & 37.44 & 30.72 & \textbf{45.8} & 38.02
    % & 49.54 & 42.34 & 38.71 & 29.5 & 43.41
    & 32.33 \\

KV² (8K)
    & \textbf{30.5} & \textbf{54.0} & 40.0 & \textbf{41.5}
    & 28.27 & 11.95 & 14.0 & 18.07
    & 43.43 & \textbf{40.97} & \textbf{31.16} & 30.2 & \textbf{38.5}
    % & -- & -- & -- & -- & --
    & 32.94 \\

KV² (32K)
    & \textbf{30.5} & 52.0 & \textbf{41.0} & 41.17
    & 28.27 & 12.87 & 16.0 & 19.05
    & 42.6 & 37.91 & 28.95 & 30.0 & 36.5
    % & 50.56 & 43.24 & 40.51 & 33.5 & 44.68
    & \textbf{33.33} \\

\bottomrule
\end{tabular*}
\end{adjustbox}

\end{table*}

%% Column to data_dir mapping for evaluation (use longbench-e):
%% Upper block:
%% NrtvQA   -> narrativeqa (no _e variant available)
%% HotpotQA -> hotpotqa_e
%% GovRep   -> gov_report_e
%% QMSum -> qmsum
%%
%% Lower block:
%% TREC     -> trec_e
%% PCount   -> passage_count_e
%% Lcc      -> lcc_e
%% RB-P     -> repobench-p_e

\section{Additional benchmark results}
This appendix reports the full, per-benchmark results that the main text only summarizes. Section~\ref{sec:appendix:ruler4k} covers RULER 4K on both backbones, Section~\ref{sec:appendix:ruler16k} gives the detailed per-backbone RULER 16K tables across the full range of KV-cache budgets, and Section~\ref{sec:appendix:infinitebench} extends the evaluation to the 200K-token InfiniteBench suite.
\label{sec:appendix:detailed-results}
\label{sec:appendix:additional-analysis}

\subsection{RULER 4K}
\label{sec:appendix:ruler4k}

\begin{figure*}[ht!]
  \centering
  \begin{minipage}[t]{0.49\linewidth}
    \centering
    \includegraphics[width=\linewidth]{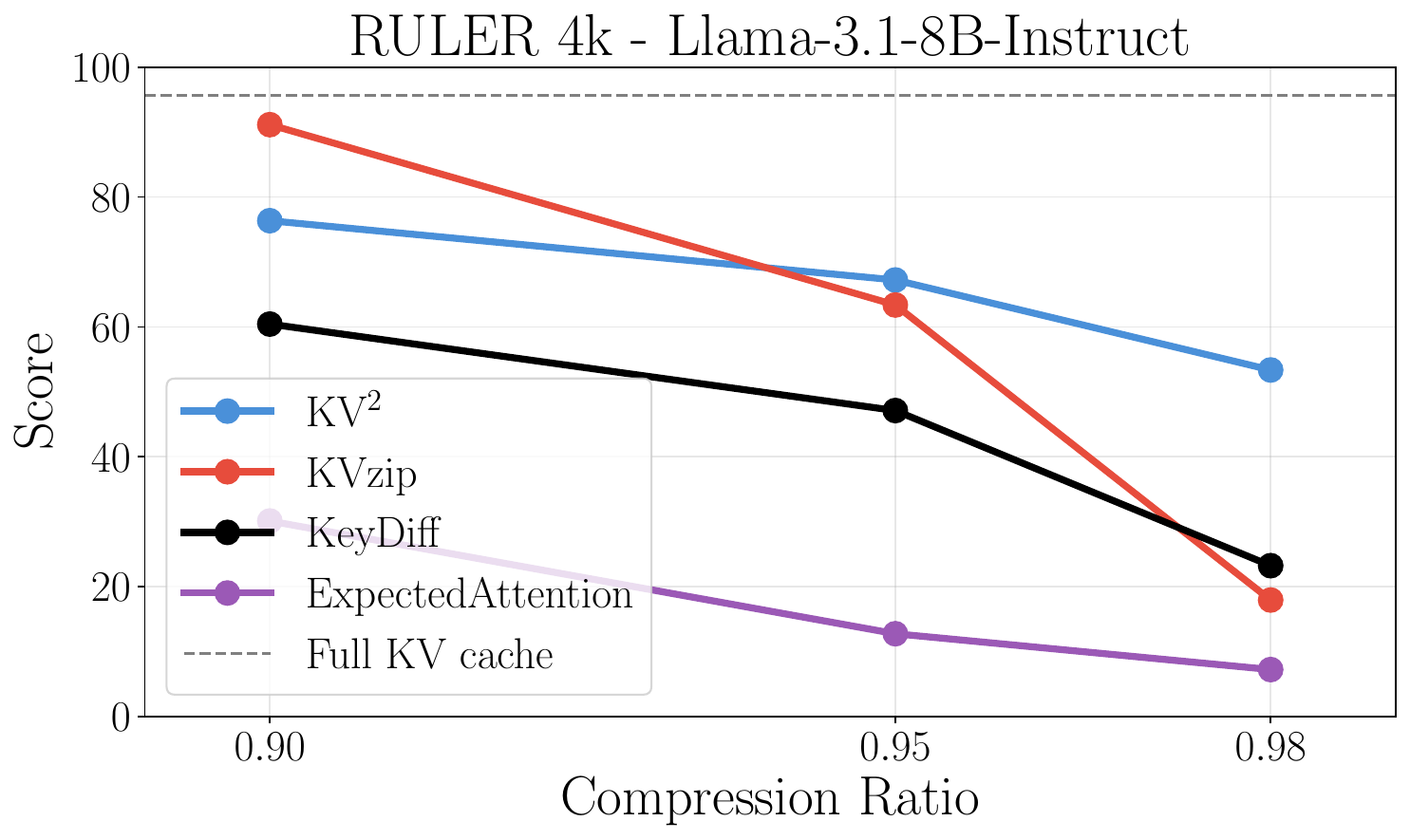}
  \end{minipage}\hfill
  \begin{minipage}[t]{0.49\linewidth}
    \centering
    \includegraphics[width=\linewidth]{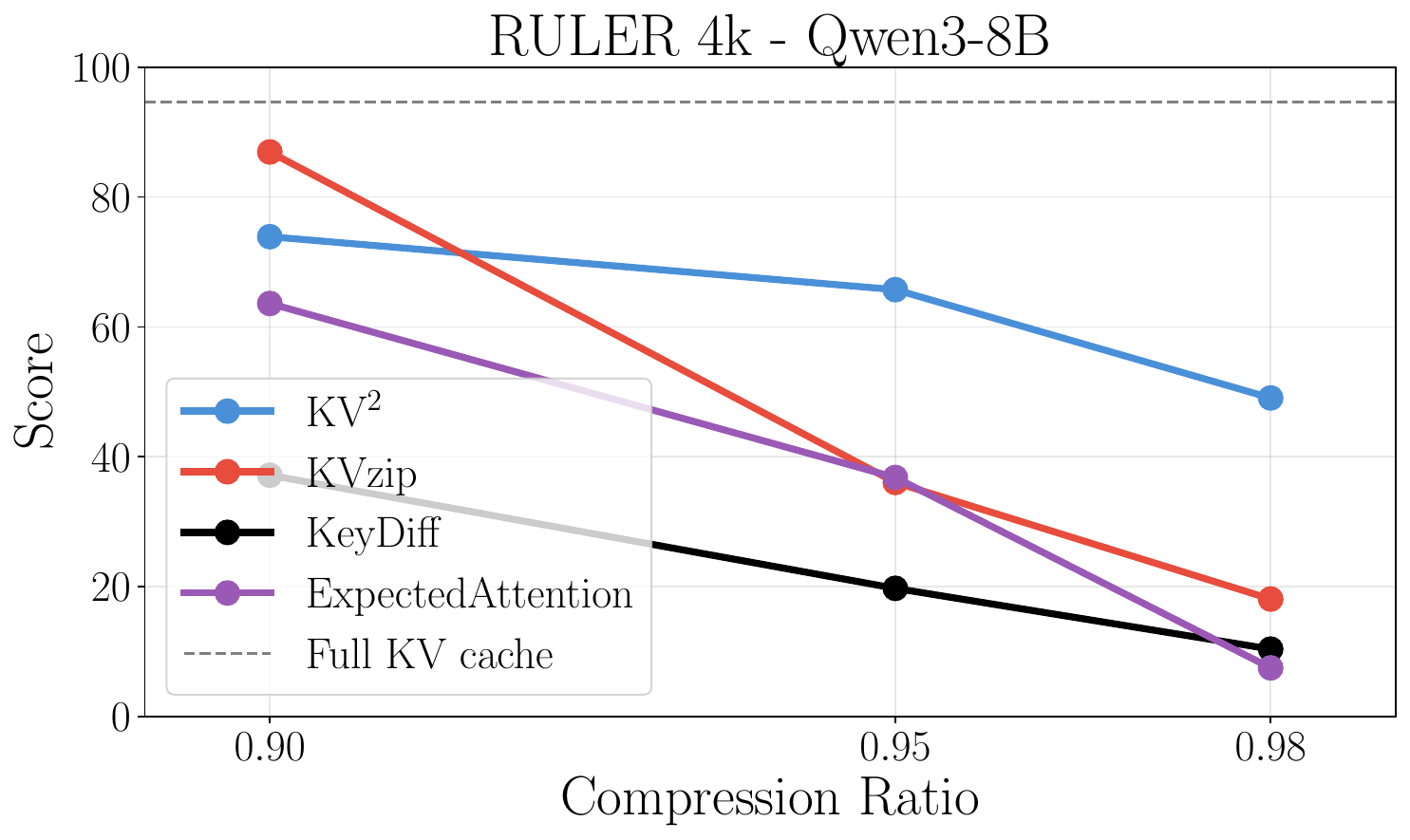}
  \end{minipage}
  \captionsetup{hypcap=false}
  \captionof{figure}[RULER 4K results across two backbones]{RULER 4K score versus KV retention ratio for \texttt{Llama-3.1-8B-Instruct} (left) and \texttt{Qwen3-8B} (right). KV$^2$ is more robust than lightweight heuristics at small budgets, while KVzip becomes more competitive as more KV entries are kept.}
  \label{fig:appendix:ruler4k}
\end{figure*}

Figure~\ref{fig:appendix:ruler4k} shows the same pattern as RULER 16K, but with a smaller gap. At 4K context length, broader reconstruction is less wasteful, so KVzip remains competitive at 10\% and 5\% KV size. Under more aggressive compression, KV$^2$ again degrades more gracefully than the lighter baselines and, especially on Qwen3-8B, than KVzip. RULER 4K therefore marks the boundary of the main result. Selective reconstruction still helps, but the advantage is naturally weaker when prompts are short enough that global reconstruction is less wasteful.

%\clearpage
\subsection{RULER 16K}
\label{sec:appendix:ruler16k}

The tables below give the full per-backbone RULER 16K results summarized in the main text, including the wider range of KV budgets.

\subsubsection{\texorpdfstring{Llama-3.1-8B-Instruct}{Llama-3.1-8B-Instruct}}
\label{sec:appendix:ruler16k-llama}
Table~\ref{tab:evaluation:ruler16k-llama} reports the complete RULER 16K results for \texttt{Llama-3.1-8B-Instruct} across every evaluated KV-cache budget, broken down by task category (aggregation, retrieval, QA, and multi-hop). Consistent with the main text, KV$^2$ preserves aggregation and retrieval accuracy far better than the baselines as the budget tightens toward 2\%.
\begin{table*}[ht!]
\centering
\thesistablestyle
\scriptsize
\setlength{\tabcolsep}{2.3pt}
\renewcommand{\arraystretch}{0.98}
\captionsetup{hypcap=false}
\captionof{table}[RULER 16K results on \texttt{Llama-3.1-8B-Instruct}]{Detailed RULER 16K results on \texttt{Llama-3.1-8B-Instruct} across a wide range of KV budgets, highlighting where KV² preserves retrieval and aggregation performance under extreme compression.}
\label{tab:evaluation:ruler16k-llama}
\begin{adjustbox}{max totalsize={\textwidth}{0.78\textheight},center}
\begin{tabular}{@{}lcccccccccccccc@{}}
\toprule
\multirow{2}{*}{\textbf{Method}}
    & \multicolumn{2}{c}{\textbf{Aggregation}}
    & \multicolumn{8}{c}{\textbf{Retrieval}}
    & \multicolumn{2}{c}{\textbf{QA}}
    & \textbf{Multi-hop}
    & \multirow{2}{*}{\textbf{Avg.}} \\
\cmidrule(lr){2-3} \cmidrule(lr){4-11} \cmidrule(lr){12-13} \cmidrule(lr){14-14}
    & cwe & fwe
    & mk\_1 & mk\_2 & mk\_3
    & mq & mv
    & s\_1 & s\_2 & s\_3
    & qa\_1 & qa\_2
    & vt
    & \\
\midrule
\multicolumn{15}{c}{\textbf{Llama-3.1-8B-Instruct}} \\
\midrule

Full
    & 89.18 & 90.2 & 99.6 & 100.0 & 99.2 & 99.0 & 99.1 & 100.0 & 100.0 & 100.0 & 81.2 & 56.8 & 99.8
    & 93.39 \\

\midrule
\multicolumn{15}{c}{\textit{KV Size = 75\%}} \\
\midrule

KeyDiff
    & 89.08 & 89.47 & 99.6 & 98.0 & 43.4 & 99.25 & 98.95 & \textbf{100.0} & \textbf{100.0} & \textbf{100.0} & 76.6 & 55.0 & 99.72
    & 88.39 \\

ExpectedAttention
    & 86.98 & 89.73 & \textbf{100.0} & 99.6 & \textbf{99.4} & 99.15 & 98.55 & \textbf{100.0} & \textbf{100.0} & \textbf{100.0} & 79.0 & 54.8 & 99.56
    & 92.83 \\

KVzip
    & 89.14 & \textbf{90.73} & 99.8 & \textbf{100.0} & 99.2 & \textbf{99.95} & 99.2 & \textbf{100.0} & \textbf{100.0} & \textbf{100.0} & 80.8 & 56.4 & 99.76
    & 93.46 \\

KV²
    & \textbf{89.42} & 90.67 & 99.8 & \textbf{100.0} & 99.2 & \textbf{99.95} & \textbf{99.25} & \textbf{100.0} & \textbf{100.0} & \textbf{100.0} & \textbf{81.0} & \textbf{56.8} & \textbf{99.8}
    & \textbf{93.53} \\

\midrule
\multicolumn{15}{c}{\textit{KV Size = 50\%}} \\
\midrule

KeyDiff
    & 87.5 & 88.8 & 99.4 & 76.0 & 6.8 & 99.35 & 99.1 & \textbf{100.0} & \textbf{100.0} & \textbf{100.0} & 65.0 & 52.8 & 99.6
    & 82.64 \\

ExpectedAttention
    & 81.1 & 88.4 & 99.6 & 96.0 & 54.8 & 99.0 & 97.65 & \textbf{100.0} & \textbf{100.0} & 77.6 & 70.8 & 54.8 & 99.48
    & 86.09 \\

KVzip
    & 88.48 & \textbf{90.67} & \textbf{99.8} & \textbf{100.0} & 99.2 & \textbf{99.9} & \textbf{99.7} & \textbf{100.0} & \textbf{100.0} & \textbf{100.0} & \textbf{82.2} & \textbf{56.2} & 99.76
    & \textbf{93.53} \\

KV²
    & \textbf{89.04} & 90.4 & 99.6 & \textbf{100.0} & \textbf{99.4} & \textbf{99.9} & 99.55 & \textbf{100.0} & \textbf{100.0} & \textbf{100.0} & 82.0 & \textbf{56.2} & \textbf{99.8}
    & \textbf{93.53} \\

\midrule
\multicolumn{15}{c}{\textit{KV Size = 25\%}} \\
\midrule

KeyDiff
    & 71.32 & 87.73 & \textbf{99.6} & 27.2 & 0.0 & 99.4 & 98.95 & \textbf{100.0} & \textbf{100.0} & \textbf{100.0} & 45.2 & 45.8 & 98.76
    & 74.92 \\

ExpectedAttention
    & 28.1 & 84.0 & 95.8 & 65.4 & 10.8 & 96.9 & 80.8 & \textbf{100.0} & 99.6 & 10.2 & 51.8 & 43.0 & 96.12
    & 66.35 \\

KVzip
    & 87.72 & 87.73 & 99.2 & \textbf{100.0} & \textbf{99.0} & 99.75 & 99.25 & \textbf{100.0} & \textbf{100.0} & \textbf{100.0} & \textbf{77.4} & \textbf{54.0} & \textbf{99.72}
    & \textbf{92.6} \\

KV²
    & \textbf{87.94} & \textbf{88.8} & 99.4 & 98.2 & 60.4 & \textbf{99.85} & \textbf{99.4} & \textbf{100.0} & \textbf{100.0} & \textbf{100.0} & 76.6 & 53.4 & \textbf{99.72}
    & 89.52 \\

\midrule
\multicolumn{15}{c}{\textit{KV Size = 10\%}} \\
\midrule

KeyDiff
    & 25.34 & \textbf{84.4} & \textbf{99.4} & 3.0 & 0.0 & 99.4 & 98.7 & \textbf{100.0} & \textbf{100.0} & \textbf{100.0} & 24.6 & 34.2 & 94.64
    & 66.44 \\

ExpectedAttention
    & 2.14 & 28.67 & 15.4 & 2.6 & 0.0 & 13.4 & 14.8 & 99.4 & 22.0 & 0.0 & 27.4 & 29.6 & 76.88
    & 25.56 \\

KVzip
    & 27.04 & 83.4 & 96.8 & \textbf{94.8} & \textbf{65.0} & 97.1 & 90.2 & \textbf{100.0} & \textbf{100.0} & \textbf{100.0} & \textbf{66.8} & \textbf{50.8} & 99.16
    & \textbf{82.39} \\

KV²
    & \textbf{74.06} & 78.67 & 99.0 & 28.4 & 0.8 & \textbf{99.8} & \textbf{99.25} & \textbf{100.0} & \textbf{100.0} & \textbf{100.0} & 57.8 & 47.2 & \textbf{99.52}
    & 75.73 \\

\midrule
\multicolumn{15}{c}{\textit{KV Size = 5\%}} \\
\midrule

KeyDiff
    & 7.96 & \textbf{83.67} & 97.0 & 1.2 & 0.0 & 98.6 & 96.3 & \textbf{100.0} & \textbf{100.0} & 99.8 & 18.0 & 28.8 & 79.64
    & 62.38 \\

ExpectedAttention
    & 0.48 & 0.87 & 0.4 & 0.0 & 0.0 & 0.4 & 1.45 & 95.4 & 0.4 & 0.0 & 23.0 & 24.8 & 34.96
    & 14.01 \\

KVzip
    & 1.22 & 47.53 & 81.4 & \textbf{66.4} & \textbf{6.6} & 69.1 & 39.65 & \textbf{100.0} & 94.2 & 86.2 & 42.0 & 33.2 & 97.44
    & 58.84 \\

KV²
    & \textbf{48.34} & 77.53 & \textbf{97.8} & 8.0 & 0.2 & \textbf{99.75} & \textbf{98.4} & \textbf{100.0} & \textbf{100.0} & \textbf{100.0} & \textbf{42.2} & \textbf{39.6} & \textbf{98.84}
    & \textbf{70.05} \\

\midrule
\multicolumn{15}{c}{\textit{KV Size = 2\%}} \\
\midrule

KeyDiff
    & 1.3 & 74.13 & 78.2 & 0.4 & \textbf{0.0} & 85.9 & 80.2 & \textbf{100.0} & \textbf{100.0} & 77.8 & 14.4 & 23.6 & 43.0
    & 52.23 \\

ExpectedAttention
    & 0.22 & 0.0 & 0.2 & 0.0 & \textbf{0.0} & 0.0 & 0.15 & 81.8 & 0.0 & 0.0 & 22.8 & 24.4 & 8.04
    & 10.59 \\

KVzip
    & 0.24 & 0.4 & 1.2 & \textbf{4.0} & \textbf{0.0} & 0.2 & 0.3 & \textbf{100.0} & 2.0 & 0.0 & 19.4 & 23.6 & 92.76
    & 18.78 \\

KV²
    & \textbf{20.22} & \textbf{77.93} & \textbf{93.2} & 0.6 & \textbf{0.0} & \textbf{98.1} & \textbf{97.9} & \textbf{100.0} & 99.6 & \textbf{93.8} & \textbf{27.0} & \textbf{31.8} & \textbf{96.72}
    & \textbf{64.37} \\

\midrule
\multicolumn{15}{c}{\textit{KV Size = 1\%}} \\
\midrule

KeyDiff
    & 0.32 & \textbf{43.6} & 31.6 & \textbf{0.2} & \textbf{0.0} & 41.85 & 40.85 & \textbf{100.0} & \textbf{99.0} & 4.4 & 14.2 & 21.4 & 10.0
    & 31.34 \\

ExpectedAttention
    & 0.18 & 0.0 & 0.0 & 0.0 & \textbf{0.0} & 0.0 & 0.0 & 48.2 & 0.0 & 0.0 & \textbf{22.2} & 25.0 & 2.0
    & 7.51 \\

KVzip
    & 0.02 & 0.0 & 0.0 & 0.0 & \textbf{0.0} & 0.0 & 0.0 & \textbf{100.0} & 0.0 & 0.0 & 20.2 & 22.6 & 88.4
    & 17.79 \\

KV²
    & \textbf{5.4} & 41.93 & \textbf{88.8} & \textbf{0.2} & \textbf{0.0} & \textbf{88.8} & \textbf{81.95} & 98.0 & 97.8 & \textbf{76.6} & 21.8 & \textbf{27.0} & \textbf{95.2}
    & \textbf{55.65} \\

\midrule
\multicolumn{15}{c}{\textit{KV Size = 0.5\%}} \\
\midrule

KeyDiff
    & 0.02 & \textbf{26.2} & 11.8 & \textbf{0.2} & \textbf{0.0} & 12.35 & 12.65 & \textbf{95.4} & 55.0 & 0.0 & 17.8 & 20.6 & 0.16
    & 19.4 \\

ExpectedAttention
    & 0.04 & 0.0 & 0.0 & 0.0 & \textbf{0.0} & 0.0 & 0.0 & 19.8 & 0.0 & 0.0 & \textbf{21.0} & 25.8 & 0.96
    & 5.2 \\

KV²
    & \textbf{0.46} & 14.2 & \textbf{68.0} & 0.0 & \textbf{0.0} & \textbf{76.25} & \textbf{67.55} & 93.4 & \textbf{93.8} & \textbf{15.6} & 19.6 & \textbf{28.6} & \textbf{94.32}
    & \textbf{43.98} \\
\bottomrule
\end{tabular}
\end{adjustbox}
\normalsize
\setlength{\tabcolsep}{6pt}
\renewcommand{\arraystretch}{1}
\end{table*}

\subsubsection{\texorpdfstring{Qwen3-8B}{Qwen3-8B}}
\label{sec:appendix:ruler16k-qwen}
Table~\ref{tab:evaluation:ruler16k} reports the corresponding RULER 16K results for \texttt{Qwen3-8B}. The same pattern holds as on Llama: the margin in favor of KV$^2$ widens under the most aggressive 2\%-5\% budgets, and is largest on the sparse retrieval and aggregation subtasks.

\begin{table*}[ht!]
\centering
\thesistablestyle
\setlength{\tabcolsep}{2.5pt}
\renewcommand{\arraystretch}{1.02}
\captionsetup{hypcap=false}
\captionof{table}[RULER 16K results on \texttt{Qwen3-8B}]{Detailed RULER 16K results on \texttt{Qwen3-8B} across a wide range of KV budgets, highlighting where KV² preserves retrieval and aggregation performance under extreme compression.}
\label{tab:evaluation:ruler16k}

\begin{adjustbox}{max totalsize={\textwidth}{0.78\textheight},center}
\begin{tabular*}{\textwidth}{@{\extracolsep{\fill}}lcccccccccccccc@{}}
\toprule
\multirow{2}{*}{\textbf{Method}}
    & \multicolumn{2}{c}{\textbf{Aggregation}}
    & \multicolumn{8}{c}{\textbf{Retrieval}}
    & \multicolumn{2}{c}{\textbf{QA}}
    & \textbf{Multi-hop}
    & \multirow{2}{*}{\textbf{Avg.}} \\
\cmidrule(lr){2-3} \cmidrule(lr){4-11} \cmidrule(lr){12-13} \cmidrule(lr){14-14}
    & cwe & fwe
    & mk\_1 & mk\_2 & mk\_3
    & mq & mv
    & s\_1 & s\_2 & s\_3
    & qa\_1 & qa\_2
    & vt
    & \\
\midrule
\multicolumn{15}{c}{\textbf{Qwen3-8B}} \\
\midrule

Full
    & 83.04 & 94.0
    & 99.6 & 100.0 & 99.6
    & 100.0 & 99.55
    & 100.0 & 100.0 & 99.8
    & 75.6 & 58.0
    & 100.0
    & 93.01 \\

\midrule
\multicolumn{15}{c}{\textit{KV Size = 75\%}} \\
\midrule

KeyDiff
    & 85.32 & \textbf{95.2}
    & 87.2 & 74.4 & 50.4
    & 98.55 & 98.85
    & \textbf{100.0} & \textbf{100.0} & \textbf{99.8}
    & 47.4 & 43.2
    & \textbf{100.0}
    & 83.1 \\

ExpectedAttention
    & \textbf{86.9} & 93.33
    & 99.0 & 99.6 & \textbf{99.8}
    & 99.95 & 99.55
    & \textbf{100.0} & \textbf{100.0} & \textbf{99.8}
    & \textbf{76.0} & 58.2
    & \textbf{100.0}
    & \textbf{93.24} \\

KVzip
    & 82.6 & 93.8
    & \textbf{99.6} & \textbf{100.0} & 99.6
    & \textbf{100.0} & \textbf{99.6}
    & \textbf{100.0} & \textbf{100.0} & \textbf{99.8}
    & 75.0 & \textbf{58.4}
    & \textbf{100.0}
    & 92.95 \\

KV²
    & 83.18 & 93.73
    & \textbf{99.6} & \textbf{100.0} & 99.6
    & \textbf{100.0} & \textbf{99.6}
    & \textbf{100.0} & \textbf{100.0} & \textbf{99.8}
    & 74.8 & 58.2
    & \textbf{100.0}
    & 92.96 \\

\midrule
\multicolumn{15}{c}{\textit{KV Size = 50\%}} \\
\midrule

KeyDiff
    & 86.24 & \textbf{94.67}
    & 82.6 & 31.6 & 14.0
    & 97.7 & 98.7
    & \textbf{100.0} & \textbf{100.0} & 99.0
    & 32.8 & 30.4
    & \textbf{100.0}
    & 74.44 \\

ExpectedAttention
    & \textbf{87.58} & 92.73
    & 99.0 & 99.6 & 99.0
    & 99.55 & 99.5
    & \textbf{100.0} & \textbf{100.0} & \textbf{100.0}
    & 71.4 & 55.8
    & \textbf{100.0}
    & 92.63 \\

KVzip
    & 83.82 & 93.53
    & \textbf{99.6} & \textbf{100.0} & 99.4
    & \textbf{100.0} & \textbf{99.7}
    & \textbf{100.0} & \textbf{100.0} & 99.8
    & \textbf{75.6} & 58.6
    & \textbf{100.0}
    & 93.08 \\

KV²
    & 83.32 & 93.6
    & \textbf{99.6} & \textbf{100.0} & \textbf{99.6}
    & 99.95 & \textbf{99.7}
    & \textbf{100.0} & \textbf{100.0} & 99.8
    & \textbf{75.6} & \textbf{59.0}
    & \textbf{100.0}
    & \textbf{93.09} \\

\midrule
\multicolumn{15}{c}{\textit{KV Size = 25\%}} \\
\midrule

KeyDiff
    & 82.92 & 88.73
    & 71.4 & 3.6 & 0.6
    & 94.25 & 97.15
    & \textbf{100.0} & 97.0 & 95.6
    & 19.6 & 18.6
    & \textbf{100.0}
    & 66.88 \\

ExpectedAttention
    & 84.18 & 92.07
    & 98.0 & 96.6 & 44.8
    & 99.55 & 99.25
    & \textbf{100.0} & 99.8 & 89.4
    & 58.2 & 48.6
    & \textbf{100.0}
    & 85.42 \\

KVzip
    & \textbf{85.08} & 92.93
    & 98.2 & \textbf{99.2} & \textbf{97.6}
    & 99.8 & 99.35
    & \textbf{100.0} & 99.8 & \textbf{100.0}
    & \textbf{74.8} & \textbf{57.6}
    & \textbf{100.0}
    & \textbf{92.64} \\

KV²
    & 84.06 & \textbf{93.47}
    & \textbf{99.4} & 99.0 & 78.0
    & \textbf{99.9} & \textbf{99.65}
    & \textbf{100.0} & \textbf{100.0} & 99.8
    & 64.6 & 55.6
    & \textbf{100.0}
    & 90.27 \\

\midrule
\multicolumn{15}{c}{\textit{KV Size = 10\%}} \\
\midrule

KeyDiff
    & 64.4 & 65.47
    & 52.4 & 0.2 & 0.0
    & 79.45 & 78.2
    & \textbf{100.0} & 69.8 & 78.2
    & 8.2 & 13.0
    & 79.88
    & 53.02 \\

ExpectedAttention
    & 35.36 & 87.13
    & 84.0 & 22.4 & 0.0
    & 94.4 & 93.9
    & 99.8 & 95.8 & 29.6
    & 39.2 & 35.2
    & 99.92
    & 62.82 \\

KVzip
    & 50.36 & \textbf{93.73}
    & 68.6 & \textbf{70.2} & \textbf{49.2}
    & 96.2 & 79.35
    & \textbf{100.0} & 98.8 & 99.6
    & \textbf{59.8} & \textbf{52.4}
    & \textbf{100.0}
    & \textbf{78.33} \\

KV²
    & \textbf{80.2} & 93.27
    & \textbf{98.0} & 40.6 & 2.6
    & \textbf{99.6} & \textbf{99.65}
    & \textbf{100.0} & \textbf{99.8} & \textbf{100.0}
    & 37.6 & 39.0
    & \textbf{100.0}
    & 76.18 \\

\midrule
\multicolumn{15}{c}{\textit{KV Size = 5\%}} \\
\midrule

KeyDiff
    & 34.74 & 44.33
    & 29.4 & 0.2 & 0.0
    & 48.4 & 48.95
    & \textbf{100.0} & 30.6 & 45.8
    & 6.6 & 10.6
    & 31.24
    & 33.14 \\

ExpectedAttention
    & 8.76 & 66.27
    & 56.6 & 0.0 & 0.0
    & 62.55 & 61.65
    & 99.8 & 81.6 & 0.4
    & 30.4 & 28.6
    & 98.36
    & 45.77 \\

KVzip
    & 3.1 & 67.67
    & 36.0 & \textbf{10.2} & \textbf{0.2}
    & 32.75 & 33.35
    & \textbf{100.0} & 26.4 & 37.8
    & \textbf{31.6} & \textbf{29.6}
    & \textbf{100.0}
    & 39.13 \\

KV²
    & \textbf{69.22} & \textbf{92.27}
    & \textbf{94.2} & 4.4 & \textbf{0.2}
    & \textbf{97.75} & \textbf{99.3}
    & \textbf{100.0} & \textbf{100.0} & \textbf{100.0}
    & 28.4 & 29.2
    & \textbf{100.0}
    & \textbf{70.38} \\

\midrule
\multicolumn{15}{c}{\textit{KV Size = 2\%}} \\
\midrule

KeyDiff
    & 1.8 & 23.33
    & 12.8 & \textbf{0.0} & \textbf{0.0}
    & 9.95 & 8.75
    & \textbf{100.0} & 5.6 & 4.6
    & 4.0 & 9.4
    & 13.36
    & 14.89 \\

ExpectedAttention
    & 2.04 & 21.33
    & 1.4 & \textbf{0.0} & \textbf{0.0}
    & 0.8 & 2.8
    & 98.4 & 2.0 & 0.0
    & \textbf{26.2} & 22.8
    & 36.72
    & 16.5 \\

KVzip
    & 0.18 & 0.0
    & 0.0 & \textbf{0.0} & \textbf{0.0}
    & 0.0 & 0.0
    & \textbf{100.0} & 0.0 & 0.0
    & 24.2 & 21.8
    & \textbf{100.0}
    & 18.94 \\

KV²
    & \textbf{23.46} & \textbf{93.53}
    & \textbf{74.0} & \textbf{0.0} & \textbf{0.0}
    & \textbf{81.45} & \textbf{90.6}
    & \textbf{100.0} & \textbf{96.8} & \textbf{100.0}
    & 23.6 & \textbf{27.2}
    & 97.8
    & \textbf{62.19} \\

\midrule
\multicolumn{15}{c}{\textit{KV Size = 1\%}} \\
\midrule

KeyDiff
    & 0.28 & 3.13
    & 1.4 & \textbf{0.0} & \textbf{0.0}
    & 1.7 & 1.8
    & \textbf{100.0} & 0.4 & 0.0
    & 4.2 & 8.8
    & 3.36
    & 9.62 \\

ExpectedAttention
    & 0.22 & 2.67
    & 0.0 & \textbf{0.0} & \textbf{0.0}
    & 0.0 & 0.0
    & 81.4 & 0.0 & 0.0
    & \textbf{26.2} & 22.8
    & 3.56
    & 10.53 \\

KVzip
    & 0.06 & 0.0
    & 0.0 & \textbf{0.0} & \textbf{0.0}
    & 0.0 & 0.0
    & \textbf{100.0} & 0.0 & 0.0
    & 23.6 & 22.4
    & \textbf{99.92}
    & 18.92 \\

KV²
    & \textbf{5.78} & \textbf{88.67}
    & \textbf{49.2} & \textbf{0.0} & \textbf{0.0}
    & \textbf{50.15} & \textbf{56.15}
    & 96.6 & \textbf{72.8} & \textbf{95.2}
    & 22.2 & \textbf{24.2}
    & 84.64
    & \textbf{49.66} \\

\midrule
\multicolumn{15}{c}{\textit{KV Size = 0.5\%}} \\
\midrule

KeyDiff
    & 0.18 & 0.13
    & 0.2 & \textbf{0.0} & \textbf{0.0}
    & 0.0 & 0.25
    & 58.6 & 0.0 & 0.0
    & 4.4 & 8.4
    & 0.68
    & 5.6 \\

ExpectedAttention
    & 0.06 & 0.13
    & 0.0 & \textbf{0.0} & \textbf{0.0}
    & 0.0 & 0.0
    & 3.4 & 0.0 & 0.0
    & \textbf{26.8} & \textbf{23.0}
    & 0.0
    & 4.11 \\

KVzip
    & 0.02 & 0.0
    & 0.0 & \textbf{0.0} & \textbf{0.0}
    & 0.0 & 0.0
    & \textbf{100.0} & 0.0 & 0.0
    & 24.4 & 22.6
    & \textbf{95.8}
    & 18.68 \\

KV²
    & \textbf{0.62} & \textbf{6.07}
    & \textbf{28.4} & \textbf{0.0} & \textbf{0.0}
    & \textbf{25.05} & \textbf{29.9}
    & 73.4 & \textbf{45.8} & \textbf{16.2}
    & 22.0 & 22.2
    & 80.08
    & \textbf{26.9} \\
\bottomrule
\end{tabular*}
\end{adjustbox}

\end{table*}

\subsection{InfiniteBench}
\label{sec:appendix:infinitebench}
Table~\ref{tab:evaluation:infinitebench} extends the comparison to InfiniteBench, whose inputs exceed 100K tokens, at 2\%, 5\%, and 10\% KV-cache budgets on a single 10\% subsample. We exclude \texttt{math\_find}, whose prompt template embeds the question, and report 8-task averages. KV$^2$ clearly leads at 2\% on \texttt{Llama-3.1-8B-Instruct} and is on par with the strongest baseline (within 0.5 points) in all other settings. Passkey and number-string retrieval remain robust under compression, whereas KV retrieval nearly collapses for all compressed methods (at most 8.00 vs.\ 60-64 with the full cache) and QA is more sensitive to reduced budgets.

\begin{table*}[ht!]
\centering
\thesistablestyle
\captionsetup{hypcap=false}
\captionof{table}[InfiniteBench benchmark results with 2\%, 5\% and 10\% KV cache sizes]{InfiniteBench benchmark results with 2\%, 5\% and 10\% KV cache sizes on a single 10\% subsample. \texttt{math\_find} is excluded because its prompt template embeds the question. Averages are over the remaining eight tasks. Passkey and number-string retrieval remain robust under compression, whereas KV retrieval nearly collapses for all compressed methods and QA is more sensitive to reduced KV budgets.}
\label{tab:evaluation:infinitebench}

\begin{adjustbox}{max totalsize={\textwidth}{0.78\textheight},center}
\begin{tabular*}{\textwidth}{@{\extracolsep{\fill}}lccccccccc@{}}
\toprule
\multirow{2}{*}{\textbf{Method}} 
    & \multicolumn{3}{c}{\textbf{Retrieval}}
    & \multicolumn{4}{c}{\textbf{QA}} 
    & \textbf{Code}
    & \multirow{2}{*}{\textbf{Avg.}} \\
\cmidrule(lr){2-4} \cmidrule(lr){5-8} \cmidrule(lr){9-9}
    & Passkey & KV-Ret & NumStr 
    & QA-Eng & QA-Chn & Choice & Dialog 
    & Debug
    & \\
\midrule

%% ============================================================
%% Llama-3-8B-Instruct
%% ============================================================
\multicolumn{10}{c}{\textbf{Llama-3.1-8B-Instruct}} \\
\midrule

Full 
    & 99.15 & 60.00 & 99.66
    & 25.92 & 31.95 & 69.43 & 21.50
    & 16.75
    & 53.05 \\

\midrule
\multicolumn{10}{c}{\textit{KV Size = 2\%}} \\
\midrule

KeyDiff 
    & 81.36 & \textbf{0.00} & 74.58
    & 14.38 & 3.86 & 21.74 & 5.00
    & \textbf{33.33}
    & 29.28 \\

KVzip 
    & \textbf{100.00} & \textbf{0.00} & 88.14
    & 9.78 & 2.00 & \textbf{34.78} & 0.00
    & \textbf{33.33}
    & 33.50 \\

KV²
    & \textbf{100.00} & \textbf{0.00} & \textbf{98.31}
    & \textbf{16.28} & \textbf{11.78} & \textbf{34.78} & \textbf{10.00}
    & \textbf{33.33}
    & \textbf{38.06} \\

\midrule
\multicolumn{10}{c}{\textit{KV Size = 5\%}} \\
\midrule

KeyDiff 
    & \textbf{100.00} & \textbf{0.00} & 84.75
    & \textbf{25.31} & 6.52 & \textbf{43.48} & \textbf{10.00}
    & \textbf{38.46}
    & 38.57 \\

KVzip 
    & \textbf{100.00} & \textbf{0.00} & 96.61
    & 14.52 & 5.99 & 34.78 & \textbf{10.00}
    & 30.77
    & 36.58 \\

KV²
    & \textbf{100.00} & \textbf{0.00} & \textbf{100.00}
    & 14.81 & \textbf{12.10} & \textbf{43.48} & \textbf{10.00}
    & 30.77
    & \textbf{38.90} \\

\midrule
\multicolumn{10}{c}{\textit{KV Size = 10\%}} \\
\midrule

KeyDiff 
    & \textbf{100.00} & \textbf{0.00} & 93.22
    & \textbf{26.37} & \textbf{12.26} & 43.48 & 5.00
    & 33.33
    & 39.21 \\

KVzip 
    & \textbf{100.00} & \textbf{0.00} & \textbf{100.00}
    & 14.06 & 4.43 & 52.17 & \textbf{25.00}
    & \textbf{41.03}
    & 42.09 \\

KV²
    & \textbf{100.00} & \textbf{0.00} & \textbf{100.00}
    & 24.34 & 7.74 & \textbf{56.52} & 15.00
    & 35.90
    & \textbf{42.44} \\

\midrule
%% ============================================================
%% Qwen2.5-7B-Instruct-1M
%% ============================================================
\multicolumn{10}{c}{\textbf{Qwen2.5-7B-Instruct-1M}} \\
\midrule

Full
    & 100.00 & 64.00 & 100.00
    & 7.66 & 2.89 & 69.57 & 15.00
    & 33.33
    & 49.06 \\

\midrule
\multicolumn{10}{c}{\textit{KV Size = 2\%}} \\
\midrule

KeyDiff
    & \textbf{100.00} & \textbf{0.00} & \textbf{100.00}
    & 5.16 & 1.05 & \textbf{39.13} & 5.00
    & \textbf{25.64}
    & 34.50 \\

KVzip
    & \textbf{100.00} & \textbf{0.00} & \textbf{100.00}
    & 3.06 & 1.14 & 17.39 & 0.00
    & 23.08
    & 30.58 \\

KV²
    & \textbf{100.00} & \textbf{0.00} & \textbf{100.00}
    & \textbf{5.46} & \textbf{1.63} & 34.78 & \textbf{20.00}
    & 17.95
    & \textbf{34.98} \\

\midrule
\multicolumn{10}{c}{\textit{KV Size = 5\%}} \\
\midrule

KeyDiff
    & \textbf{100.00} & \textbf{0.00} & \textbf{100.00}
    & \textbf{6.72} & 1.19 & \textbf{65.22} & 5.00
    & 15.38
    & \textbf{36.69} \\

KVzip
    & \textbf{100.00} & \textbf{0.00} & \textbf{100.00}
    & 5.52 & 1.57 & 43.48 & 5.00
    & 23.08
    & 34.83 \\

KV²
    & \textbf{100.00} & \textbf{0.00} & \textbf{100.00}
    & 5.92 & \textbf{2.70} & 47.83 & \textbf{10.00}
    & \textbf{25.64}
    & 36.51 \\

\midrule
\multicolumn{10}{c}{\textit{KV Size = 10\%}} \\
\midrule

KeyDiff
    & \textbf{100.00} & 0.00 & \textbf{100.00}
    & 7.23 & 1.47 & \textbf{73.91} & 5.00
    & 20.51
    & 38.52 \\

KVzip
    & \textbf{100.00} & \textbf{8.00} & \textbf{100.00}
    & \textbf{7.58} & \textbf{2.96} & 65.22 & 10.00
    & \textbf{33.33}
    & 40.89 \\

KV²
    & \textbf{100.00} & 0.00 & \textbf{100.00}
    & 5.51 & 2.83 & \textbf{73.91} & \textbf{20.00}
    & 25.64
    & \textbf{40.99} \\
\bottomrule
\end{tabular*}
\end{adjustbox}

\end{table*}

%% ============================================================
%% Column to data_dir mapping for evaluation:
%% ============================================================
%% Passkey  -> passkey
%% KV-Ret   -> kv_retrieval
%% NumStr   -> number_string
%% QA-Eng   -> longbook_qa_eng
%% QA-Chn   -> longbook_qa_chn
%% Choice   -> longbook_choice_eng
%% Dialog   -> longdialogue_qa_eng
%% Sum      -> longbook_sum_eng
%% Debug    -> code_debug
%% Find     -> math_find (excluded: its prompt template embeds the question)
%% ============================================================

%\clearpage

\section{Token-level reconstruction example}
\label{sec:appendix:selection-example}
Figure~\ref{fig:appendix:selection-example} gives a qualitative view of which tokens the first-stage proxy selects for reconstruction. It shows a condensed RULER variable-tracking prompt with the positions chosen by the KeyDiff proxy (after averaging over layers and KV heads) highlighted. The selected tokens concentrate on the instruction, the variable-assignment chain, and the question-answer span, whereas the repetitive filler passages contribute only a sparse scattering of tokens, illustrating the signal sparsity that selective reconstruction is designed to exploit.

\begin{figure*}[ht!]
  \centering
  \setlength{\fboxsep}{2pt}
  \fbox{%
  \begin{minipage}{0.96\linewidth}
  \raggedright
  \footnotesize\ttfamily
  \detokenize{<|begin_of_text|><|start_header_id|>system<|end_header_id|>}\\[0.2ex]
  Cutting Knowledge Date: December 2023\\
  Today Date: 26 Jul 2024\\[0.4ex]
  \detokenize{<|eot_id|><|start_header_id|>user<|end_header_id|>}\\[0.2ex]
  \selectionhighlight{Memorize and track the chain(s) of variable assignment hidden in the following text.}\\
  \selectionhighlight{The grass is green. The sky is blue. The sun is yellow. Here we go.} There \selectionhighlight{and back again.}\\
  \selectionhighlight{VAR QPE = 64886} The grass \selectionhighlight{is} green. \selectionhighlight{The} sky is blue. The sun is yellow. Here we go. There and back again.\\
  \selectionhighlight{The} grass is green. The sky is blue. The sun is yellow. Here \selectionhighlight{we} go. There \selectionhighlight{and} back again.\\
  \selectionhighlight{VAR SEJ = VAR QPE} \selectionhighlight{The} grass \selectionhighlight{is} green. \selectionhighlight{The} sky is blue. The sun is yellow. Here \selectionhighlight{we} go. There and back again.\\
  \selectionhighlight{VAR ZQO = VAR SE}J The grass is green. The sky is blue. The sun is yellow. Here we go. There and back again.\\
  VAR \selectionhighlight{RVU = VAR Z}QO The grass is green. The sky is blue. The sun is yellow. Here \selectionhighlight{we} go. There and back again.\\
  The grass is green. The sky is blue. The sun is yellow. Here we go. There and back again.\\
  The grass is green. The sky is blue. The sun is yellow. Here we go. There and back again.\\
  The grass is green. The sky is blue. The sun is yellow. Here we go. There and back again.\\
  \selectionhighlight{VAR FAI = VAR RV}U The grass is green. The sky is blue. The sun is yellow. Here we go. There and back again.\\
  \selectionhighlight{Question: Find all variables that are assigned the value 64886 in the text}\\
  Answer: According to the \selectionhighlight{chain(s) of variable assignment in the text above, 5 variables are assgined}\\
  the value 64886, they are: \selectionhighlight{QPE} \selectionhighlight{SEJ} ZQO \selectionhighlight{RV}U \selectionhighlight{FAI}\\[0.8ex]
  \selectionhighlight{Memorize and track the chain(s) of variable assignment hidden in the following text.}\\[0.3ex]
  \selectionhighlight{The} grass is green. The sky is blue. The sun is yellow. Here we go. There and back again.\\
  ...\\
  \selectionhighlight{VAR RHCUJ = 98223} The grass is green. The sky is blue. The sun is yellow. Here \selectionhighlight{we} go. There and back again.
  \end{minipage}}
  \captionsetup{hypcap=false}
  \captionof{figure}[Qualitative reconstruction-token selection]{Condensed variable-assignment prompt with the selected tokens highlighted. The highlights reflect the first-stage KeyDiff proxy after averaging over layers and KV heads and then selecting the top token positions for reconstruction. Some highlights cover only partial words because the selection is shown at token granularity rather than word granularity. The dominant mass falls on the instruction, assignment chain, and question-answer span, with only sparse selections from the repetitive continuation.}
  \label{fig:appendix:selection-example}
\end{figure*}

\end{document}